\documentclass[a4paper,fleqn]{cas-dc}

\usepackage{hyperref}
\usepackage[numbers, square, comma, sort&compress]{natbib}
\usepackage[normalem]{ulem} 

\usepackage{empheq} 
\usepackage{xcolor}
\usepackage{ulem}
\usepackage{siunitx}
\usepackage{extdash}
\usepackage{comment}
\usepackage{soul}

\usepackage{bbm}
\usepackage{caption}
\usepackage{subcaption}

\newcommand{\LD}{\color{blue}}

\newcommand{\BS}{\color{magenta}}
\newcommand{\cnj}[1]{{\color{red}#1}}

\def\tsc#1{\csdef{#1}{\textsc{\lowercase{#1}}\xspace}}
\tsc{WGM}
\tsc{QE}
\tsc{EP}
\tsc{PMS}
\tsc{BEC}
\tsc{DE}
\usepackage{amsthm}
\newtheorem{remark}{Remark}

\newcommand{\newText}[1]{\textcolor{black}{#1}}
\newcommand{\remText}[1]{\iffalse {#1} \fi}

\newcommand{\nText}[1]{\textcolor{black}{#1}}
\newcommand{\rText}[1]{\iffalse {#1} \fi}

\usepackage{nomencl} 
\usepackage{framed} 
\usepackage{multicol} 
\makenomenclature
\renewcommand*\nompreamble{\begin{multicols}{2}}
\renewcommand*\nompostamble{\end{multicols}}

\begin{document}
\let\WriteBookmarks\relax
\def\floatpagepagefraction{1}
\def\textpagefraction{.001}
\shorttitle{Physically Consistent Neural Networks for building thermal modeling: theory and analysis}
\shortauthors{Di Natale et~al.}


\title [mode = title]{Physically Consistent Neural Networks for building thermal modeling: theory and analysis}




\author[1,2]{Di Natale L.}[orcid=0000-0002-3295-412X]
\cormark[1]
\credit{Conceptualization, Methodology, Software, Validation, Formal analysis, Data Curation, Visualization, Writing - Original Draft}

\author[1]{Svetozarevic B.}
\credit{Conceptualization, Methodology, Writing - Review \& Editing,  Supervision }

\author[1]{Heer P.}
\credit{Writing - Review \& Editing, Resources, Funding acquisition}

\author[2]{Jones C.N.}
\credit{Conceptualization, Methodology, Writing - Review \& Editing, Supervision}

\address[1]{Urban Energy Systems Laboratory, Swiss Federal Laboratories for Materials Science and Technology (Empa), 8600 D\"{u}bendorf, Switzerland}
\address[2]{Laboratoire d'Automatique, Swiss Federal Institute of Technology Lausanne (EPFL), 1015 Lausanne, Switzerland}

\cortext[cor1]{Corresponding author:  \texttt{loris.dinatale@empa.ch} (L. Di Natale)}


\begin{abstract}
Due to their high energy intensity, buildings play a major role in the current worldwide energy transition. Building models are ubiquitous since they are needed at each stage of the life of buildings, i.e. for design, retrofitting, and control operations. Classical white-box models, based on physical equations, are bound to follow the laws of physics but the specific design of their underlying structure might hinder their expressiveness and hence their accuracy. On the other hand, black-box models are better suited to capture nonlinear building dynamics and thus can often achieve better accuracy, but they require a lot of data and might not follow the laws of physics, a problem that is particularly common for neural network (NN) models. 
To counter this known generalization issue, physics-informed NNs have recently been introduced, where researchers 
introduce prior knowledge in the structure of NNs to ground them in known underlying physical laws and avoid classical NN generalization issues. 

In this work, we present a novel physics-informed NN architecture, dubbed Physically Consistent NN (PCNN), which only requires past operational data and no engineering overhead, including prior knowledge in a linear module running in parallel to a classical NN. We formally prove that such networks are physically consistent -- by design and even on unseen data -- with respect to different control inputs and temperatures outside and in neighboring zones. We 
demonstrate their performance on a case study, where the PCNN attains an accuracy up to \newText{$40\%$}\remText{$50\%$} better than a classical physics-based resistance-capacitance model on $3$-day long prediction horizons. Furthermore, despite their constrained structure, PCNNs attain similar performance to classical NNs on the validation data, overfitting the training data less and retaining high expressiveness to tackle the generalization issue.  
\end{abstract}

\begin{keywords}
Neural Networks \\
Physical consistency \\
Prior knowledge \\
Building models \\
Deep Learning

\end{keywords}


\maketitle
\makenomenclature

\begin{table*}[!t]   

\begin{framed}
\nomenclature[01]{\textit{PCNN variables}}{}
\nomenclature[02]{$D$}{Unforced dynamics}
\nomenclature[03]{$E$}{Energy accumulator}
\nomenclature[04]{$P$}{Power}
\nomenclature[05]{$Q^{sun}$}{Solar gains}
\nomenclature[06]{$T$}{Temperature of the modeled zone}
\nomenclature[07]{$T^{neigh}$}{Temperature of the neighboring zone}
\nomenclature[08]{$T^{out}$}{Outside temperature}
\nomenclature[09]{$T^{w}$}{Water temperature of the heating system}
\nomenclature[10]{$a$}{Heating effect scaling parameter}
\nomenclature[11]{$b$}{Heat losses to the outside scaling parameter}
\nomenclature[12]{$c$}{Heat losses to the neighboring zone scaling parameter}
\nomenclature[13]{$d$}{Cooling effect scaling parameter}
\nomenclature[14]{$\dot m$}{Water mass flow rate in a radiator}
\nomenclature[15]{$u$}{Control inputs}
\nomenclature[16]{$x$}{Inputs to the black-box module}
\nomenclature[17]{\textit{Grey-box model variables}}{}
\nomenclature[18]{$\xi$}{Disturbance model}
\nomenclature[19]{$u$}{Controllable inputs}
\nomenclature[20]{$w^1, w^2$}{Uncontrollable inputs}
\nomenclature[21]{$z$}{State of the system}

\printnomenclature

\end{framed}

\end{table*}

    \section{Introduction}
    \label{sec: introduction}
    
Buildings consume $30\%$ of global end-use energy, producing $28\%$ of the world's Green House Gas (GHG) emissions related to energy according to the IEA \cite{iea2020buildings}, and those proportions rise to $40\%$ of the total energy usage and $36\%$ of the total GHG in the European Union (EU) \cite{ec2019tfactsheet}. Space heating and cooling have a major impact, with heating alone being responsible for $64\%$ of household energy consumption in the EU \cite{eurostat2020households}. To follow the Paris Agreement pledges to limit global warming to well below \SI{2}{\celsius} \cite{unfccc_paris_2015}, there is thus a need to decrease the energy intensity of the building sector.

    \subsection{The importance of building models}
    
There are three technology-driven ways to attain this decarbonization objective: through better designs, retrofit\nText{s}\rText{ting}, or improved operations of buildings. In all cases, models play a central role, either to find the best design \cite{westermann2019surrogate}, the most effective refurbishment \cite{rabani2020minimizing}, or to learn intelligent controllers to replace poorly performing rule-based controllers \cite{svetozarevic2021data}. 

Modeling buildings is a challenging task in general since inside temperatures, air quality, or visual comfort, among others, all depend on highly stochastic exogenous 
factors mainly driven by the weather and the behavior of the occupants \cite{boodi2018intelligent, fan2017short, shamsi2020framework, tian2018review}. Additionally, the introduction of solar panels, heat pumps, battery storage, electric vehicles, 
and other new technologies makes it harder to model building operations as a whole and calls for scalable and flexible methods \cite{shamsi2021feature}.

Most of the existing models are focused on commercial buildings \cite{boodi2018intelligent} and study single-step predictors \cite{afroz2018modeling, cai2019day}, which work adequately for energy consumption predictions in design or retrofitting applications.
On the other hand, in this work, we design control-oriented thermal models for a residential case study that could, for example, be used in 
\nText{to learn}\rText{a} Reinforcement Learning (RL) control \nText{policies}\rText{strategy}. This calls for short-term multi-step temperature predictions, to be able to minimize energy consumption while maintaining the comfort of the occupants. Indeed, while RL can generally be applied in a model-free fashion, \rText{learning via interactions with the real system,} 
the data-inefficiency of RL algorithms \cite{wang2020reinforcement} and the slow dynamics of buildings often require agents to be trained over thousands of days of data \rText{to attain sufficient performance} \cite{wan2018residential, yu2019deep}\nText{, which is not feasible in practice.}\rText{. This is, however, not feasible in practice and we still rely on simulations to train RL agents.} 

While classically engineered physics-based models still dominate the field, researchers recently started to leverage the growing amount of available data to design data-driven building models. Such models generally perform better, are more flexible, and rely on less technical knowledge, but require a lot of past data to be trained on and lack generalization guarantees outside of the training data \cite{afroz2018modeling}. This is particularly true for models based on Neural Networks (NNs), which can be very data-inefficient and fail 
when new input\nText{s they} \rText{data the networks}were not trained on \rText{is}\nText{are} fed to them~\cite{djeumou2021neural}, \nText{which is known as their \textit{generalization issue}}. In particular, there are no guarantees that a classical NN \rText{captures}\nText{follows} the underlying physics, \rText{i.e. follows}\nText{e.g.} the laws of thermodynamics in the case of thermal modeling\rText{, which}\nText{. This} is however critical for control-oriented applications\rText{, as detailed below.}\nText{, e.g. to ensure the RL agents trained on these models capture the impact of heating and cooling correctly.}
    
\rText{The nomenclature has been added.}    

    \subsection{The generalization issue of neural networks}

Originally spotted by Szegedy et al. \cite{szegedy2013intriguing}, the generalization issue of NNs led to the field of adversarial examples, where researchers aim to find \rText{minimal} input perturbations that fool NNs, showing how brittle their predictions can be~\cite{wiyatno2019adversarial, moosavi2016deepfool}\nText{, even when only little noise is applied to the input}. \rText{For example, DeepFool selects an image and applies a little noise -- indistinguishable to the human eye -- that changes the prediction of a well-trained NN \cite{moosavi2016deepfool}.}

To circumvent this generalization issue, researchers often rely on better sets of data that cover the entire spectrum of inputs and allow NNs to react to any situation. This requires vast amounts of resources and is only possible in fields where a significant amount of data is available, such as for tasks related to natural language processing \cite{brown2020language} or images \cite{xie2020self}. Additionally, to ensure some level of generalization, practitioners typically separate the data into training and validation sets, the former being used to train the network and the latter to assess its performance on unseen data \nText{to avoid \textit{overfitting} the training data} \cite{xu2018splitting}.\rText{ This ensures the NN presents at least some generalization capability and avoids \textit{overfitting} the training data.} However, classical NNs cannot be robust to input modifications that 
do not exist in the entire data set.

\begin{figure*}
    \begin{center}
    \includegraphics[width=\textwidth]{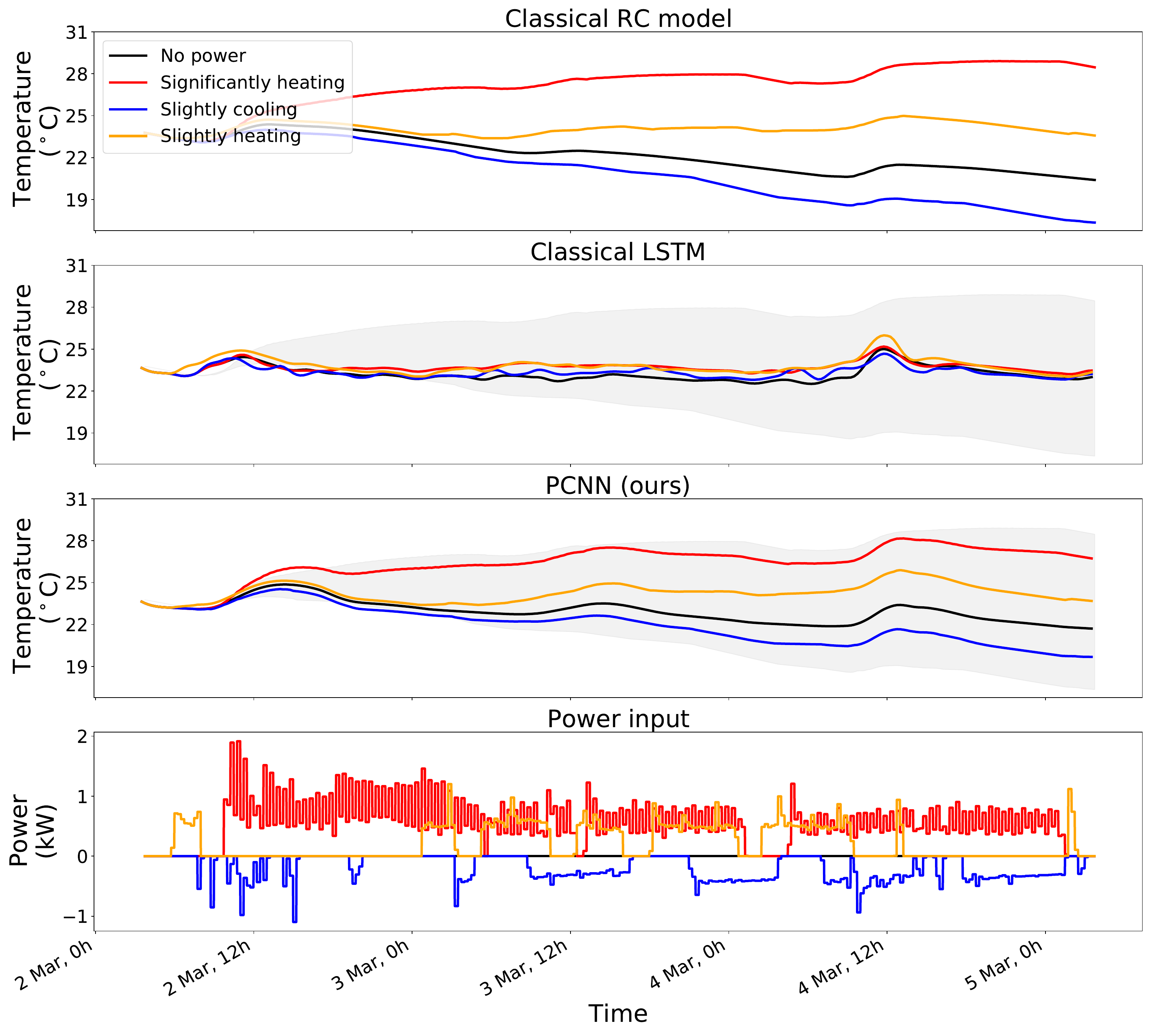}
    \caption{Temperature predictions of the RC model and the proposed PCNN detailed and analyzed in Section~\ref{sec: results} compared to a classical LSTM under different control inputs. The grey-shaded areas represent the span of the RC model predictions to provide a visual comparison with both black-box methods. While the LSTM presents a lower training error than the PCNN (see Section~\ref{sec: results}), indicating a good fit to the data, it does not capture the impact of the different heating/cooling powers applied to the system, 
    e.g. predicting higher temperatures when cooling is on than when heating is. The specific structure of PCNNs introduced in Section~\ref{sec: methods}, on the other hand, allows them to retain physical consistency, similarly to classical physics-based models, while improving the prediction accuracy (see Section~\ref{sec: accuracy}).}
    \label{fig: inconsistency}
    \end{center}
    \end{figure*}

In the case of building thermal models, even if several years of data are available, one will always face an input coverage problem. Indeed, buildings are usually inhabited and operated in a typical fashion to maintain a comfortable temperature -- heating when it gets cold in winter and cooling when it gets hot in summer. Most data sets are hence inherently incomplete and we cannot hope to learn robust NNs that grasp the effect of heating in summer, for example.
When predicting the evolution of the temperature over long horizons of several days, classical NNs might therefore fail to capture the underlying physics, i.e. the impact of heating and cooling on the temperature\rText{, as }\nText{. This is illustrated}\rText{pictured} in Figure~\ref{fig: inconsistency}\rText{. In this plot}, \nText{ where} one can compare the temperature predictions of a classical physics-based resistance-capacitance (RC) model, a classical Long Short-Term Memory network (LSTM), and a Physically Consistent NN (PCNN) proposed in this work under different heating and cooling power inputs. Interestingly, the LSTM achieves a superior accuracy than both other models on the training data, overfitting it, but clearly fails to capture the impact of heating and cooling. 

    \subsection{Introducing physics-based prior knowledge}
    
In general, classical NNs 
suffer from \textit{underspecification}, as reported in a large-scale study from Google \cite{d2020underspecification}. As a countermeasure, we should find ways to include prior knowledge, typically about the underlying laws of physics, into NNs to facilitate their training and improve their performance. This trend already began several years ago with the emergence of physics-guided machine learning \cite{karpatne2017physics} and the creation of specific network structures that represent known physical systems \cite{lutter2019deep, greydanus2019hamiltonian, djeumou2021neural}. In such NNs, physics can, for example, be introduced directly in the structure of the network or through custom loss functions, among others \cite{von2019informed}. In this paper, we refer to these \newText{models}\remText{works} as Physics-informed Neural Networks (PiNNs)\newText{.} \remText{and we propose a new architecture, dubbed PCNN, applied to thermal building modeling.} 

\newText{To the best of the authors' knowledge, Drgo\v{n}a et al. \cite{drgona2021physics} were the first to use PiNNs as control-oriented building models, but they did not provide theoretical guarantees of their models following the underlying physics, except for the hard-encoded dissipativity. Furthermore, the performance of their models, which remarkably work in the multi-zone setting, was not benchmarked against classical methods. Concurrently to our work, Gokhale et al. developed another PiNN structure for control-oriented building modeling, but they modified the loss function of their NNs and not their architecture \cite{gokhale2021physics}, contrary to PCNNs. Finally, while not relying on PiNNs, we want to mention here the recent work of B\"{u}nning et al. on physics-inspired linear regression for buildings \cite{bunning2021physics}, which is philosophically related to the general efforts to introduce physical priors in otherwise black-box models.}\remText{To the best of the authors' knowledge, Drgo\v{n}a et al. were the first to use PiNNs as control-oriented building models \cite{drgona2021physics}. They replaced the state, input, disturbance, and output matrices of classical linear models with four NNs and leveraged known physical rules to enforce constraints on them. They additionally used the Perron-Frobenius theorem to enforce the stability and dissipativity of the system by bounding the eigenvalues of all the NNs. However, they did not provide any theoretical guarantees of their model following the underlying laws of physics in general beyond the hard-encoded dissipativity. Furthermore, the performance of their model, which remarkably works in the multi-zone setting, was not contrasted with a baseline to ensure it performs well despite the additional structural constraints introduced. 
Concurrently to our work, Gokhale et al. developed another PiNN structure for control-oriented building modeling \cite{gokhale2021physics}. However, their approach relies on the introduction of a new physics-inspired loss term to guide the learning of the NN towards physically meaningful solutions rather than a modification of the NN architecture, contrary to PCNNs. While not relying on PiNNs, the recent work of Bünning et al. on physics-inspired linear regression for buildings \cite{bunning2021physics} is philosophically related to the general efforts to introduce physical priors in otherwise black-box models.}


\remText{\\
Note: The physical consistency definition was moved to the methods, Section 3.1.}

    \subsection{Contribution}
    
\rText{In this work, t}\nText{T}o tackle the aforementioned generalization issues of classical NNs, we introduce a novel PiNN architecture, dubbed PCNN, which includes existing knowledge on the physics of the system at its core, with an application to building zone temperature modeling. The introduction of prior knowledge \rText{in the models} essentially works as an inductive bias, such that \rText{the models}\nText{PCNNs} do not need to learn everything from data, but only what we cannot easily characterize a priori.

While PCNNs model unforced temperature dynamics\footnote{Throughout this work, \textit{unforced dynamics} represent the temperature evolution in the zone when no heating or cooling is applied and heat losses are neglected.} with classical NNs, they treat parts of the inputs separately: the power input to the zone and the heat losses to the environment and neighboring zones are processed in parallel by a linear module inspired \rText{from}\nText{by} classical physics-based RC models. This module \nText{ensures the positive correlation between power inputs and zone temperatures}\rText{allows us to control the impact of various control inputs} 
while forcing heat losses 
to be proportional to the corresponding temperature gradients\rText{, which ensures} \newText{\rText{we get physically consistent predictions}}\nText{ to provide physically consistent predictions}\remText{the conditions in Equations~(\ref{equ: consistency power})-(\ref{equ: consistency neigh})}. This solves parts of the generalization issue of NN building models and makes PCNNs well-suited for control applications. The key however is that, unlike in classical physics-based models, no engineering effort is required to design and identify the parameters of PCNNs: we only need access to past data, and PCNNs are then trained in an end-to-end fashion to learn all the parameters simultaneously. Furthermore, we show that PCNNs achieve better accuracy than a baseline RC model on a case study. Moreover, they attain a precision on par with classical LSTMs on the validation data, despite performing worse on the training data. This shows that PCNNs do not lose much expressiveness due to their constrained architecture 
and have less tendency to overfit the training data.
The main contributions of this work can be summarized as follows:
\begin{itemize}
    \item PCNNs, novel \nText{PiNNs}\rText{physics-informed neural networks}, are introduced and applied to zone temperature modeling.
    \item The physical consistency of \rText{the }PCNN\nText{s} \rText{architecture }with respect to control inputs and exogenous temperatures\footnote{\nText{We refer to the temperature outside and in neighboring zones together as \textit{exogenous temperatures} in this work.}} is formally proven.
    \item PCNNs \nText{perform comparably to LSTMs on the validation data and overfit the training data less.}\rText{ do not overfit the training data as much as classical LSTMs while achieving similar accuracy on the validation data.}
    \item PCNNs attain better performance than classical RC models on a case study while avoiding any engineering overhead.

\end{itemize}

The rest of the paper is structured as follows. We start with a brief overview of the building modeling and PiNN literature in Section~\ref{sec: background}\remText{ and of the case study in Section~\ref{sec: case study}}. We then \newText{describe what it means to be \textit{physically consistent with respect to a given input},} present the PCNN architecture and formally prove its physical consistency in Section~\ref{sec: methods}\nText{. The case study is described in Section~\ref{sec: case study}, and}\rText{. and describe the case study in Section~\ref{sec: case study}.} 
\rText{In Section~\ref{sec: results},} 
we \nText{then} compare the performance of this method against a classical RC model and LSTMs and provide a graphical \rText{and empirical }interpretation 
of its physical consistency \nText{in Section~\ref{sec: results}}. 
Finally, we discuss the potential and limitations of PCNNs in Section~\ref{sec: discussion} \nText{and}\rText{while} Section~\ref{sec: conclusion} concludes the paper.

    \section{Background}
    \label{sec: background}
    
\begin{figure}
    \begin{center}
    \includegraphics[page=1,width=\columnwidth]{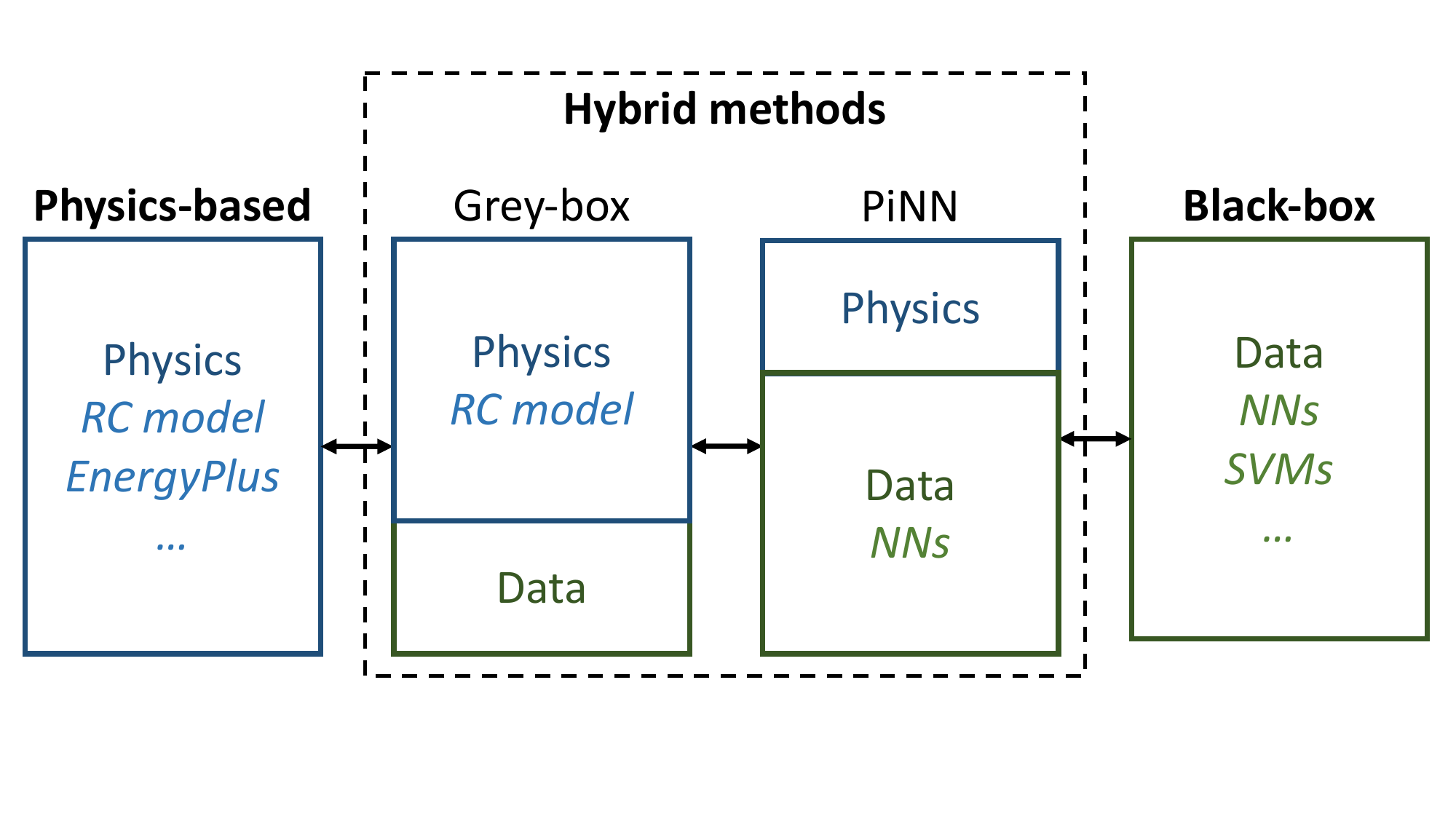}
    \caption{Structural differences between the \nText{different methods}\rText{ types of model}.}
    \label{fig: paradigm}
    \end{center}
    \end{figure}

This section presents an overview of the existing literature on building models, which can be broadly classified into three categories: physics-based, black-box, and hybrid methods, as pictured in Figure~\ref{fig: paradigm}. While the latter can generally be further broken down into grey-box approaches and PiNNs, only grey-box modeling was previously applied to buildings, with the exception of \cite{drgona2021physics, gokhale2021physics}. For this reason, we propose a more general overview of PiNNs in Section~\ref{sec: pinns}. 

Due to the vast literature on building modeling, we only provide a short summary of the strengths and weaknesses of the different techniques, and more details can be found in dedicated reviews, such as \cite{homod2013review, foucquier2013state, li2014review, deb2017review, boodi2018intelligent, afroz2018modeling, bourdeau2019modeling, ali2021review}.
    


    \subsection{Physics-based building models}
    \label{sec: physics-based}
    
Also known as white-box or first principle models, physics-based models rely on Ordinary Differential Equations (ODEs), such as convection, radiation, or conduction equations, to describe building thermal dynamics. These methods were dominating the field early on when the lack of available data hindered the development of data-driven models \cite{homod2013review}. 

Since they are grounded in first principles, a natural advantage of these approaches is the interpretability of the solutions \cite{homod2013review}. Additionally, this gives them interesting generalization capabilities outside of the training data \cite{deb2017review}. On the other hand, however, due to the complexity of detailed thermal models, assumptions and simplifications have to be made, such as in the choice of the ODEs, which can limit the accuracy of physics-based models \cite{fan2017short}. Moreover, the more precision desired, the more knowledge and time is required to design the model and find the corresponding parameters, typically concerning the building envelope and the HVAC system, which might introduce uncertainty \cite{wei2019deep, tian2018review}. 


To simplify the design of physics-based models, various detailed simulation tools were developed, such as EnergyPlus, Modelica, TRNSYS, or IDA ICE \cite{crawley2001energyplus, wetter2006modelica, mazzeo2020energyplus}. While these models can attain good accuracy and respect the underlying physical laws, they are notoriously hard to calibrate \cite{ding2019octopus, zhang2019whole, chakrabarty2021scalable}, entail considerable development and implementation costs to find and detail all the required parameters \cite{harb2016development}, and suffer from a high computational burden at run-time \cite{ascione2017artificial}.
    
    \subsection{Black-box building models}
    \label{sec: data-driven}
    
As opposed to physics-based methods, black-box (or data-driven) models do not rely on first principles but derive patterns from historical operational data. 
The most widely used methods rely on Multiple Linear Regression (MLR), Support Vector Regression (SVR), NNs, and ensembles, apart from the classical Autoregressive Integrated Moving Average (ARIMA) models, as reviewed by Bourdeau et al. \cite{bourdeau2019modeling}.

Black-box models are generally easier to use than physics-based ones since no expert knowledge is required at the design stage, 
but often lack generalization guarantees outside of the data they are trained on \cite{homod2013review, afroz2018modeling}. Furthermore, they need historical data as input, sometimes in large amounts, to achieve satisfactory accuracy \cite{afroz2018modeling}, and the data additionally has to be \textit{exciting enough}, i.e. to cover the different operating conditions of the building, something not trivial, as discussed in Section~\ref{sec: introduction}. The subsequent data imbalance issue can, for example, be tackled through the creation of sub-models, like in Zhang et al. \cite{zhang2021problem}. Moreover, black-box models are sensitive to the choice of features -- or feature extraction methods -- used as model inputs \cite{fan2017short}. On the other hand, an advantage of data-driven methods is their flexibility, as they can be scaled to large systems in a more straightforward manner than physics-based methods \cite{royer2016towards}. \rText{Similarly}\nText{Additionally}, they are generally easier to transfer from one building to another since similar model architectures can be used and \nText{all }the parameters are learned from \rText{the }data. 

Very recently, as a consequence of the growing amount of available data, Deep Learning (DL) has started to be applied to building modeling \cite{bourdeau2019modeling}. For example, recurrent NNs (RNNs) were shown to provide better accuracy than feedforward NNs for the prediction of energy consumption \cite{rahman2018predicting}. In another study, a specific gated Convolutional NN (CNN) was shown to outperform RNNs and the classical Seasonal ARIMAX model on day-ahead multistep hourly predictions of the electricity consumption \cite{cai2019day}. Due to the nonconvexity of classical NN-based models, which makes them hard to use in optimization procedures, researchers also used specific control-oriented models, such as Input Convex NN (ICNN), to model building dynamics \cite{bunning2021input}.

    \subsection{Hybrid methods}
    \label{sec: hybrid}
    
Hybrid methods combine physics-based knowledge with existing data to have the best of both worlds. Note that some researchers use the term “hybrid methods” to refer to the fact that they first build a physics-based model and then fit a black-box model to it to then accelerate the inference procedure at run-time, such as \cite{ascione2017artificial, li2021modelling}, which is out of the scope of this overview and hence not covered here. 

    \subsubsection{Grey-box building models}
    \label{sec: grey-box}

In grey-box modeling, one generally starts from simplified physics-based equations and uses data-driven methods to identify the model parameters \cite{shamsi2020framework, shamsi2021feature, harb2016development} and/or learn an unknown disturbance model on top of it \cite{gray2018hybrid}. 
The simplified base model requires less expert knowledge and time to be designed than pure physics-based models but still allows one to retain the interpretability of physics-based models. Furthermore, this basis includes physical knowledge in the model, so that \rText{fewer parameters have}\nText{less information has} to be learned \nText{from data }compared to pure black-box models, which in turn implies that less historical data is required to fit such models \cite{foucquier2013state}. \rText{However, even when they are identified from given data, there is still uncertainty around the parameters \cite{shamsi2020framework}.}



\rText{Typically, people }\nText{Typical grey-box models }start with linear state-space models and identify their parameters \rText{with}\nText{from} data, even if some nonlinearities are not well captured by this approach \cite{royer2016towards}. Due to the \rText{complex parameter identification procedure}\nText{difficulty of finding good parameters in general}, low complexity RC models usually perform better\rText{, see for example \cite{fux2014ekf, berthou2014development, harb2016development} and the references therein}, with models with \nText{one or }two capacitances usually being selected \cite{fux2014ekf, berthou2014development, harb2016development}. Higher-order models \nText{furthermore} entail more complexity and hinder the generalization capability of grey-box models, which also advocates in favor of low-complexity frameworks~\cite{shamsi2019generalization}. \rText{Consequently}\nText{As a partial solution}, a feature assessment framework to test the flexibility, scalability, and interoperability of grey-box models and select the right model characteristics was proposed by Shamsi et al. \cite{shamsi2021feature}. In essence, grey-box approaches hence allow for a trade-off between the accuracy and \nText{the }complexity of \rText{the}\nText{building} models~\cite{shamsi2019generalization}.

\nText{Due to the effectiveness of low-order RC models, we hence rely on linear first-order RC modeling techniques inspired from B\"{u}nning et al. \cite{bunning2022physics} and simplified versions of Maasoumy et al. \cite{maasoumy2011model, maasoumy2014handling, maasoumy2014selecting} to construct the PCNNs proposed in this work.}

    \subsubsection{Physics-informed neural networks}
    \label{sec: pinns}


While early DL applications used classical feedforward NNs, researchers soon realized how transferring prior knowledge to NNs could be beneficial. Among the success stories, one can find the CNN and RNN families, specially designed to capture spatial invariance \cite{kayhan2020translation} and temporal dependencies \cite{yu2019review} in the data, respectively. 

In recent years, a new field emerged in the Machine Learning community to tackle the generalization issue of neural networks and create new NN architectures bound to follow given physical laws, such as Hamiltonian NNs \cite{greydanus2019hamiltonian} or Lagrangian NNs \cite{lutter2019deep}, later generalized by Djeumou et al. \cite{djeumou2021neural}. In parallel, \rText{researchers started to look at various ways to include physics-based knowledge in DL frameworks, supported by the underspecification plaguing NNs \cite{d2020underspecification}. The field was pioneered in 2017 by}\nText{PiNN architectures flourished, pioneered by }the physics-guided NNs of Karpatne et al. \cite{karpatne2017physics, karpatne2017theory} and the more general physics-informed Deep Learning (DL) framework \nText{originally} proposed by Raissi et al. \cite{raissi2017physics, raissi2017physics2, yang2018physics} \rText{and the corresponding stochastic extension \cite{yang2018physics}}. Since then, various methods to include prior knowledge in NNs have been proposed, several of which can be found in \cite{von2019informed}, where the authors tried to classify them. 

Methodologically, the PCNNs proposed in this work are close to the physics-interpretable shallow NNs, where the inputs are also processed by two modules in parallel, one to retain physical exactness when possible and one to capture nonlinearities through a shallow NN \cite{yuanphysics}. Also related in spirit to the PCNN architecture, Hu et al. introduced a specific learning pipeline, where the output of the forward NN is fed back through a physics-inspired NN  structure to reconstruct the input and hence ensure the forward process retains physical consistency \cite{hu2020physics}. \remText{Finally, as mentioned, Drgo\v{n}a et al. \cite{drgona2021physics} and Gokhale et al. \cite{gokhale2021physics} are the only ones that already applied PiNNs to thermal building modeling.} 

\newText{Finally, two recent works applied PiNNs to create control-oriented building models \cite{drgona2021physics, gokhale2021physics}.  Drgo\v{n}a et al. replaced the state, input, disturbance, and output matrices of classical linear models with four NNs and leveraged known physical rules to enforce constraints on them \cite{drgona2021physics}. They additionally used the Perron-Frobenius theorem to enforce the stability and dissipativity of the system by bounding the eigenvalues of all the NNs. 
On the other hand, Gokhale et al. relied on a more classical PiNN approach with the introduction of a new physics-inspired loss term to guide the learning \rText{of the NN }towards physically meaningful solutions without modifying the NN architecture \cite{gokhale2021physics}. However, neither of these works provide physical consistency guarantees, unlike \rText{PCNNs}\nText{the PCNN architecture presented in this work}.}

\remText{\\
Note: The case study section was moved after the methods, it is now Section 4.}

    \section{Methods}
    \label{sec: methods}
    
This section \newText{firstly defines a notion of physical consistency and} \nText{then }details the novel PCNN structure proposed in this work, where the effect of the control inputs and the heat losses to the environment and neighboring zones are separated from the unforced temperature dynamics. \rText{We then}\nText{Finally, we} formally prove \newText{the physical consistency of PCNNs}\remText{its physical consistency} with respect to control inputs and exogenous temperatures\rText{[The footnote on exogenous temperatures was moved to 1.4]}.

\remText{\\
Note: This is a new Section, the physical consistency was orginally defined in the introduction.}

    \subsection{\rText{Physical consistency}\nText{Respecting the underlying physical laws}}

Throughout this work, we define a model as being \textit{physically consistent with respect to a given input} when any change in this input leads to a change of the output that follows the underlying physical laws. 
In our case, for example, we need models that are physically consistent with respect to control inputs to ensure that turning the heating on leads to higher zone temperatures than when heating is off, and vice versa for cooling. Mathematically, we can express this requirement as follows \nText{for a zone with}\rText{, for} power input\rText{s} \nText{$P\in\mathbb{R}$} at time step $j$ and temperature prediction\rText{s} \nText{$T\in\mathbb{R}$} at time step $k$:
\begin{align}
    \frac{\partial T_k}{\partial P_j} > 0 \qquad \forall 0\leq j<k,  \label{equ: consistency power}
\end{align}
where we consider $P<0$ in the cooling case by convention throughout this paper. 
We can similarly define physical consistency with respect to the outside temperature \rText{$T^{out}$}\nText{$T^{out}\in\mathbb{R}$}, the temperature in a neighboring zone \rText{$T^{neigh}$}\nText{$T^{neigh}\in\mathbb{R}$}, and the solar gains \rText{$Q^{sun}$}\nText{$Q^{sun}\in\mathbb{R}$} as follows:
\begin{align}
    \frac{\partial T_k}{\partial T^{out}_j} > 0 \qquad \forall 0\leq j<k, \label{equ: consistency out} \\
    \frac{\partial T_k}{\partial T^{neigh}_j} > 0 \qquad \forall 0\leq j<k, \label{equ: consistency neigh} \\
    \frac{\partial T_k}{\partial Q^{sun}_j} > 0 \qquad \forall 0\leq j<k, \label{equ: consistency sun}
\end{align}
since higher exogenous temperatures or solar gains all lead to increased zone temperatures.

    \subsection{Physically consistent neural networks}
    \label{sec: black_box}
    
    \begin{figure*}
    \begin{center}
    \includegraphics[page=1,width=0.95\textwidth]{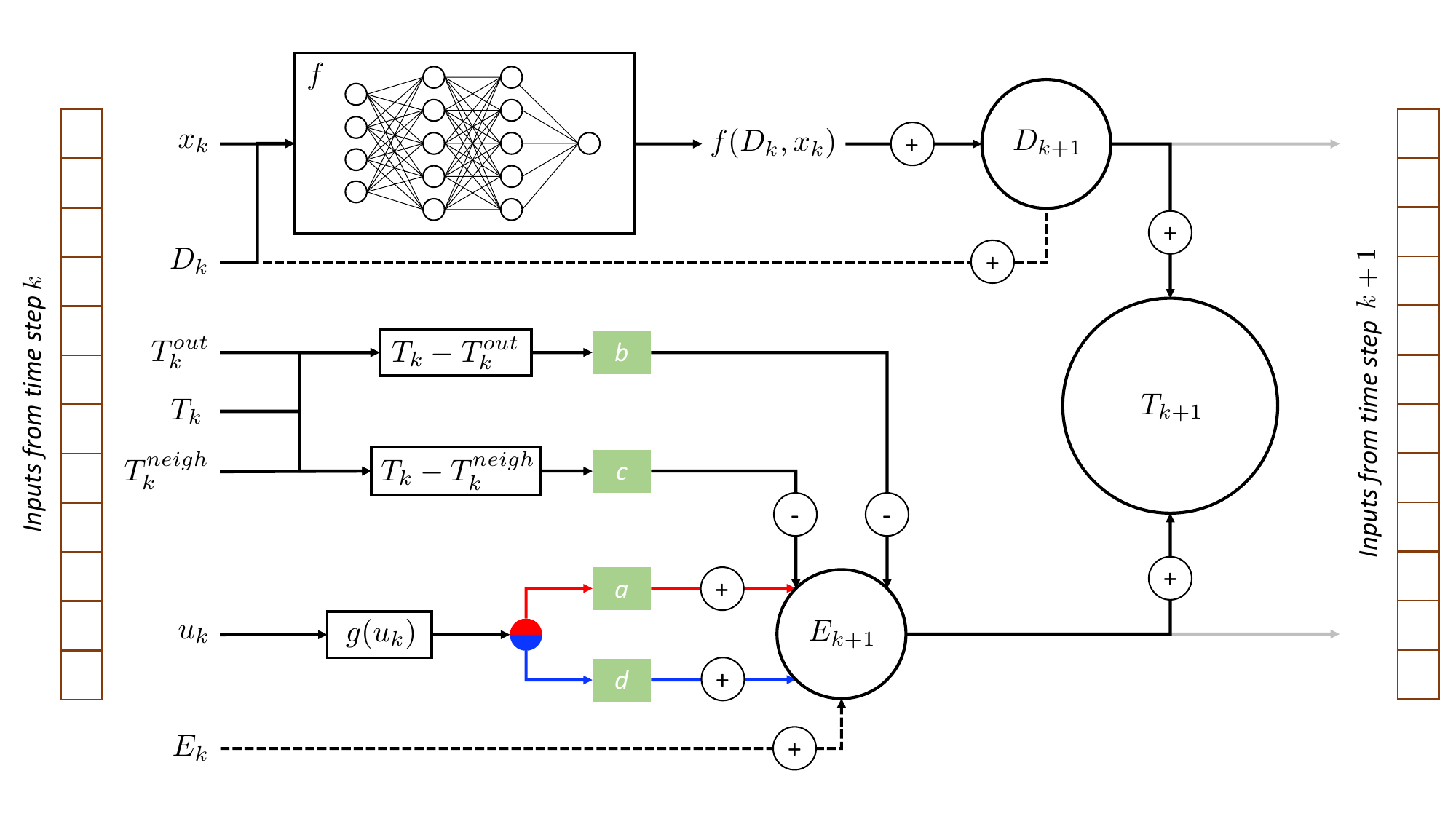}
    \caption{\rText{P}\nText{The proposed} PCNN architecture used recursively at each time step. The control inputs $u$, transformed in\nText{to} power inputs by the function $g$, and the losses to the environment $b(T-T^{out})$ and neighboring zone $c(T-T^{neigh})$ all influence an energy accumulator $E$, which accumulate\nText{s} or dissipates energy at each time step. Here, the separation between red and blue lines signals a different treatment of the power inputs in the heating and cooling case, respectively, since they are scaled by different constants $a$ and $d$. The accumulated energy is then added to the unforced dynamics $D$, modeled by a residual NN that takes all the features apart from $u$, $T^{out}$, and $T^{neigh}$ -- gathered in $x$ -- as input, to get the final zone temperature prediction $T$.}
    \label{fig: structure}
    \end{center}
    \end{figure*}
    
The \nText{proposed} PCNN architecture is sketched in Figure~\ref{fig: structure} for one time step $k$, and we apply it recursively over the prediction horizon. The temperature of the zone $T$ is \rText{predicted}\nText{computed} 
as the sum of two latent variables evolving through time: the \textit{unforced dynamics} \nText{$D\in\mathbb{R}$}, and the \rText{energy accumulated in the }\textit{energy accumulator} \nText{$E\in\mathbb{R}$}, which includes prior knowledge about thermal dynamics. \nText{Mathematically, we thus have:}
\begin{align}
    \nText{T_{k+1}} &= \nText{D_{k+1} + E_{k+1}} \label{equ: T=D+E}
\end{align}
\begin{remark}[\nText{Extension to several neighboring zones}]
\nText{While we describe and analyze the case with a single neighboring zone throughout this paper, it is straightforward to extend PCNNs to model a zone connected to several other zones by adding further energy loss terms with their corresponding scaling constants.}
\end{remark}

    \subsubsection{\nText{Linear physics-inspired module}}
    
\rText{Indeed,} \nText{The energy accumulator} $E$ is firstly positively influenced by the power input to the \rText{room}\nText{zone} \nText{$g(u)\in\mathbb{R}$}, \rText{computed}\nText{which depends on the control input $u\in\mathbb{R}^m$, e.g. the opening pattern of radiator valves.}\rText{by $g$, a function that transforms the control input $u$ 
in power input into the zone.} The latter is scaled by a constant $a$ in the heating and $d$ in the cooling case to represent its effect on the air mass in the room. Note that the power input is negative in the cooling case by convention, so that cooling lowers the energy accumulated in $E$, as expected. \rText{Additionally, the energy accumulator is losing} 

\nText{Secondly, from the laws of thermodynamics, we know that the modeled zone loses} energy through heat transfer\nText{s} to the environment and the neighboring zone. We hence subtract these effects, which are proportional to \nText{the corresponding temperature gradients with the outside temperature \nText{$T^{out}$}, respectively the temperature in the neighboring zone \nText{$T^{neigh}$}}\rText{difference between the internal temperature and the outside one $T^{out}$, respectively the one in the neighboring zone $T^{neigh}$}, scaled by parameters $b$, respectively $c$\nText{, learned from data}.\remText{Note that the current zone temperature is obtained again as the sum of both latent variables.} 
\nText{Mathematically, in the heating case, we can hence write the evolution of the physics-inspired module as follows:}
\begin{equation}
    \nText{E_{k+1}} = \nText{E_k + a g(u_k) - b (T_k - T^{out}_k) - c (T_k - T^{neigh}_k),} \label{equ: energy accumulator}
\end{equation}
\nText{with $E_0=0$. In the cooling case, one simply needs to exchange the parameter $a$ with $d$. 
As can readily be seen, 
Equation~\eqref{equ: energy accumulator} is heavily inspired by classical first-order RC building models, found e.g. in B\"{u}nning et al. \cite{bunning2022physics} and detailed in Appendix~\ref{app: general rc}. The main difference with the generic RC model in Equation~\eqref{equ: physics-based 2} is that the proposed PCNN architecture allows us to treat nonlinear solar and additional unknown heat gains using neural networks or other nonlinear functions in $D$ instead of relying on engineered linear solutions. Furthermore, the physics-inspired parameters $a$, $b$, $c$, and $d$ are learned from data simultaneously to the parameters of the black-box module described below (see Section~\ref{sec: training}).}

\begin{remark}[\nText{Design of $g$}] \label{rem: g}
\nText{In some cases, we can directly control the heating or cooling power input to the zone, i.e. $g(u)=u$. When this is not possible, e.g. when $u$ controls the opening of the valves in radiators, we need 
to process the controllable inputs into power inputs through some function $g$. 
This function might be engineered, for example as $g(u) = u*\dot m*(T^{w} - T)$ in the case of a radiator, with $\dot m$ the mass flow and $T^{w}$ the temperature of the water in the pipes, or it could be learned from data, e.g. using NNs. This learned function should be strictly monotonically increasing with $g(0)=0$, i.e. no energy is consumed when there is no control input, $g(u)<0$ when cooling is on, and $g(u)>0$ when heating is applied. 
Importantly, since everything is trained together in an end-to-end fashion (see Section~\ref{sec: training}), $g$ can seamlessly be learned in parallel to the other parameters.}
\end{remark}

\begin{remark}[\nText{Coupling between $D$ and $E$}]
\nText{Note that since $T_k = D_k + E_k$, the nonlinear black-box module $D$ influences the evolution of the energy accumulator $E$ in Equation~\eqref{equ: energy accumulator}, which is one of the main differences with classical grey-box techniques, where the physics-based and black-box modules are usually completely separated. This furthermore requires learning the parameters $a$, $b$, $c$, and $d$ simultaneously to the NNs in $D$, as presented in Section~\ref{sec: training}. Further details on the differences between classical grey-box approaches and PCNNs are discussed in Section~\ref{sec: contrast}.}
\end{remark}

    \subsubsection{\nText{Black-box module}}
    
\rText{Concurrently, $D$ is evolving as a residual NN using all the inputs that are not processed in $E$, such as solar gains and time information, gathered in $x$, to capture the temperature evolution when no heating or cooling is applied and heat losses are neglected.} \nText{Running in parallel of the linear module, the nonlinear black-box module processes all inputs not treated in $E$, such as solar gains and time information, gathered in $x\in\mathbb{R}^n$, to capture the unforced temperature dynamics, i.e. when no heating or cooling is applied and heat losses are neglected. This can typically be modeled using residual NNs, which leads to the following expression:}
\begin{align}
            \nText{D_{k+1}} &= \nText{D_k + f(D_k, x_k)} \label{equ: unforced dynamics} \\
            \nText{D_0} &= \nText{T(t_0),} \nonumber
\end{align}
\rText{Both latent variables are then carried on to the next time step, the unforced dynamics are again processed by $f$ and energy keeps accumulating in or dissipating from $E$. As can readily be seen, the physics-informed module $E$ is heavily inspired by RC models, such as the one presented in Appendix~\ref{app: general rc}. However, in PCNNs, $a$, $b$, $c$, and $d$ are learned from data simultaneously to the parameters of the NNs in $f$ using backpropagation through time (BPTT).} 
\nText{where $T(t_0)$ is the measured temperature at the beginning of the prediction horizon and $f$ is a potentially highly nonlinear function, typically based on recurrent neural networks.} Remarkably, the unforced dynamics $D$ are independent of power inputs and heat losses by design\nText{, which will allow us to prove the physical consistency of PCNNs with respect to these inputs in Section~\ref{sec: theory}}. \rText{They are processed at each time step by a potentially highly nonlinear function $f$ to capture the dynamics not represented in $E$. The key, however, is that the control inputs and exogenous temperatures do not pass through $f$, 
which will allow us to prove the physical consistency of PCNNs with respect to these inputs in Section~\ref{sec: theory}.} \rText{However and importantly, PCNNs do not require any engineering or knowledge about the building structure or parameters beyond connectivity information, i.e. which zones are adjacent, to remain consistent.}

\rText{While we picture and describe the case with a unique neighboring zone here, it is straightforward to extend PCNNs to model a zone with multiple neighbors 
by adding further energy loss terms with their corresponding scaling constants.} \rText{Furthermore, w}


\begin{remark}[\nText{Design of $f$}]
\nText{While $f$ is composed of an encoder-LSTM-decoder structure in our case (see \newText{Section~\ref{sec: implementation}}\remText{Appendix~\ref{app: results}}), any NN architecture -- and even functions that do not contain NNs -- can be used without affecting the physical consistency of the predictions. Nonetheless, due to the sequential nature of temperature dynamics and the expressiveness of NNs, we suspect RNNs to be a good choice in general.}
\end{remark}

    \subsubsection{\nText{Training procedure}}
    \label{sec: training}

\nText{Importantly, PCNNs do not require any engineering or knowledge about the building structure or parameters beyond connectivity information, i.e. which zones are adjacent to the modeled one.} 
\nText{Given a data set of measurements $\mathcal{D} = \{\tilde x(t), T(t)\}_{t=1}^{N}$, where $\tilde x = \{x,u,T^{out},T^{neigh}\}$ and $N$ is the number of data points, we can directly optimize all the parameters of both the physics-inspired and black-box modules together.}

\nText{To that end, we first construct a set of $S$ time series $\mathcal{D}_S = \{\tilde x^{(s)}(t), T^{(s)}(t)\}_{s=1}^{S}$, each consisting of consecutive data points from $\mathcal{D}$. Note that these sequences might overlap in practice to increase the data efficiency of the proposed method. We then minimize the Mean Squared Error (MSE) between the PCNN predictions $T_{k+1}(\tilde x^{(s)}_{0:k})$, recursively computed from past inputs $\tilde x^{(s)}_{0:k}$ up to time $k$, and the true measurements $T^{(s)}(k+1)$ for each time series $s$ over the predefined prediction horizon $H$:}
\begin{align}
    \frac{1}{S}\sum_{s=1}^{S}\left[{\frac{1}{H}\sum_{k=0}^{H-1}{\left(T_{k+1}(\tilde x^{(s)}_{0:k}) - T^{(s)}(k+1)\right)^2}}\right],
\end{align}
\nText{In this work, we parametrize $f$ using RNNs and rely on the standard automatic Backpropagation Through Time (BPTT) algorithm~\cite{werbos1990backpropagation} and the PyTorch library~\cite{NEURIPS2019_9015} to solve this optimization problem. \newpage}

\rText{Lastly, on the design of $g$: in some cases, we can directly control the heating or cooling power input to the zone, i.e. $g(u)=u$. When this is not possible, e.g. when $u$ controls the opening of the valves in radiators, we need 
to process the controllable inputs into power inputs through some function $g$. 
This function might be engineered, e.g. $g(u) = u*\dot m*(T^{water} - T)$ in the case of a radiator, with $\dot m$ the mass flow and $T^{water}$ the water temperature in the pipes, or it could be learned from data, for example with neural networks. This learned function should be strictly monotonically increasing with $g(0)=0$, i.e. no energy is consumed when there is no control input, $g(u)<0$ when cooling is on, and $g(u)>0$ when heating is applied, and can then be included in the pipeline in Figure~\ref{fig: structure}. Importantly, since everything is learned together in an end-to-end fashion, $g$ can seamlessly be learned in parallel to the other parameters.}

    \subsection{\rText{Physical consistency by design}\nText{PCNNs follow physical laws by design}}
    \label{sec: theory}
    
\rText{In the heating case, the temperature predictions of the PCNNs pictured in Figure~\ref{fig: structure} at time step $k+1$ can be described mathematically as follows:
\begin{align}
    D_{k+1} &= D_k + f(D_k, x_k) \\
    E_{k+1} &= E_k + a g(u_k) - b (T_k - T^{out}_k) \nonumber \\
            &\quad - c (T_k - T^{neigh}_k) \\ 
    T_{k+1} &= D_{k+1} + E_{k+1} 
\end{align}
Since we are ultimately interested in the temperature prediction, we can rewrite it as:
\begin{align}
    T_{k+1} &= T_k + f(D_k, x_k) + a g(u_k) \nonumber \\
            &\quad - b (T_k - T^{out}_k) - c (T_k - T^{neigh}_k) \label{equ: PCNN}
\end{align}
In the cooling case, one simply needs to exchange the parameter $a$ with $d$.} 
\nText{Plugging Equations~\eqref{equ: energy accumulator}-\eqref{equ: unforced dynamics} in Equation~\eqref{equ: T=D+E}, we get:}
\begin{align}
    \nText{T_{k+1}} &= \nText{T_k + f(D_k, x_k) + a g(u_k)} \nonumber \\
            &\quad\nText{ - b (T_k - T^{out}_k) - c (T_k - T^{neigh}_k)} \label{equ: PCNN}
\end{align}
\rText{Due to the resemblance between the energy accumulator and a classical RC model, Equation~(\ref{equ: PCNN}) can almost be seen as a grey-box model of the thermal dynamics.
However, there are two main differences between PCNNs and classical grey-box models, as discussed in Section~\ref{sec: contrast}: the unforced "\newText{main}\remText{base}" dynamics are captured by a black-box function, and all the parameters are learned in parallel in an end-to-end fashion.}

\noindent Applying Equation~(\ref{equ: PCNN}) recursively, as detailed in Appendix~\ref{proof: black-box}, one can express the temperature prediction of the PCNNs at any future time step $i$ as follows:
\begin{align}
    T_{k+i} &= (1-b-c)^i T_k \nonumber \\
            &\quad + \sum_{j=1}^{i}(1-b-c)^{(j-1)} {[f(D_{k+i-j}, x_{k+i-j})} \label{equ: final black-box} \\
            &\quad \qquad + {a g(u_{k+i-j}) + b T^{out}_{k+i-j} + c T^{neigh}_{k+i-j}} ] \nonumber
\end{align}
Here, it is important to note that $D_{m+1} = D_{m} + f(D_m, x_m)$ is independent of the variables $u$, $T^{out}$, and $T^{neigh}$ \nText{at any step $m$}, it solely depends on \nText{the other inputs in} $x$, so we do not need to explicitly write the recursion out.

In the case when we can directly control the power input to the zone, i.e. $g(u) = u$, we can now formally assess the physical consistency of PCNNs with respect to control inputs and exogenous temperatures since 
we get the following partial derivatives:
\begin{align}
    \frac{\partial T_{k+i}}{\partial u_{k+i-j}} &= (1-b-c)^{(j-1)}a \qquad \forall j=1,..,i. \label{equ: partial derivative 1}\\
    \frac{\partial T_{k+i}}{\partial T^{out}_{k+i-j}} &= (1-b-c)^{(j-1)}b \qquad \forall j=1,..,i. \\
    \frac{\partial T_{k+i}}{\partial T^{neigh}_{k+i-j}} &= (1-b-c)^{(j-1)}c \qquad \forall j=1,..,i. \label{equ: partial derivative 2}
\end{align}
Remarkably, these derivatives take the same form in a classical RC model, as shown in Equation~(\ref{equ: final rc}), Appendix~\ref{proof: physics-based}. 
PCNNs hence satisfy the physical consistency criteria of Equations~(\ref{equ: consistency power})-(\ref{equ: consistency neigh}) as long as the conditions below hold:
\begin{align}
    a,b,c &> 0 \nonumber \\
    1-b-c &> 0 \label{equ: conditions}
\end{align}
This is the case for real systems since $a$, $b$, and $c$ are small positive physical constants, i.e. inverses of resistances and capacitances. Moreover, it gives us simple verification criteria to ensure that the learned PCNN stays physically consistent as these conditions could easily be enforced during the training of the models, even though it was not needed in our experiments\footnote{The values learned by PCNNs in practice have orders of magnitude $10^{-1}$-$10^{-2}$ for $a$ and $d$ and $10^{-3}$-$10^{-4}$ for $b$ and $c$.}. 

Note that even when we do not have access to the power input directly and have to process the control inputs through an engineered or learned function $g$, we still get:
\begin{align}
    \frac{\partial T_{k+i}}{\partial g(u_{k+i-j})} &= (1-b-c)^{(j-1)}a \qquad \forall j=1,..,i,
\end{align}
which remains positive under the same conditions. 
This ensures that any change in the power input, as computed by a function $g$, still yields the expected physically consistent outcome on the zone temperature. As long as $g$ \rText{is constructed in a physically consistent manner }\nText{satisfies the conditions in Remark~\ref{rem: g}}\rText{, i.e. it is a strictly monotonically increasing function of $u$ with a root at $u=0$, as explained at the end of Section~\ref{sec: black_box}}, we furthermore observe that:
\rText{
\begin{align}
    \frac{\partial T_{k+i}}{\partial u_{k+i-j}} &= \frac{\partial T_{k+i}}{\partial g(u_{k+i-j})}\frac{\partial g(u_{k+i-j})}{\partial u_{k+i-j}} \nonumber \\
    &= (1-b-c)^{(j-1)}a \frac{\partial g(u_{k+i-j})}{\partial u_{k+i-j}} \label{equ: up to cst} \\
    &= (1-b-c)^{(j-1)}a h(u_{k+i-j}), \nonumber
\end{align}}
\begin{align}
    \nText{\frac{\partial T_{k+i}}{\partial u_{k+i-j}}} &= \nText{\frac{\partial T_{k+i}}{\partial g(u_{k+i-j})}\frac{\partial g(u_{k+i-j})}{\partial u_{k+i-j}}} \nonumber \\
    &= \nText{(1-b-c)^{(j-1)}a \frac{\partial g(u_{k+i-j})}{\partial u_{k+i-j}}}, \label{equ: up to cst}
\end{align}
\nText{which remains positive, hence satisfying Equation~(\ref{equ: consistency power}), as long as the conditions in Equation~(\ref{equ: conditions}) hold since $g$ is defined as a monotonically increasing function.}
\rText{where $h(u)$ is a positive function since $g$ is monotonically increasing in $u$. The latter condition ensures that the partial derivatives of the temperature predictions with respect to control inputs remain positive, hence satisfying Equation~(\ref{equ: consistency power}), as long as the conditions in Equation~(\ref{equ: conditions}) hold.}



\remText{\\
Note: The implementation details are now described in the case study, Section 4.2.}

\begin{remark}[\nText{Condition on $d$}]
\nText{Replacing $a$ by $d$ throughout Equations~\eqref{equ: PCNN}-\eqref{equ: up to cst} yields similar conditions for the cooling case. In particular, we require to have $d>0$ to retain physical consistency, additionally to the conditions in Equation~\eqref{equ: conditions}.}
\end{remark}

    \section{Case study}
    \label{sec: case study}

\begin{figure}
    \begin{center}
    \includegraphics[width=\columnwidth]{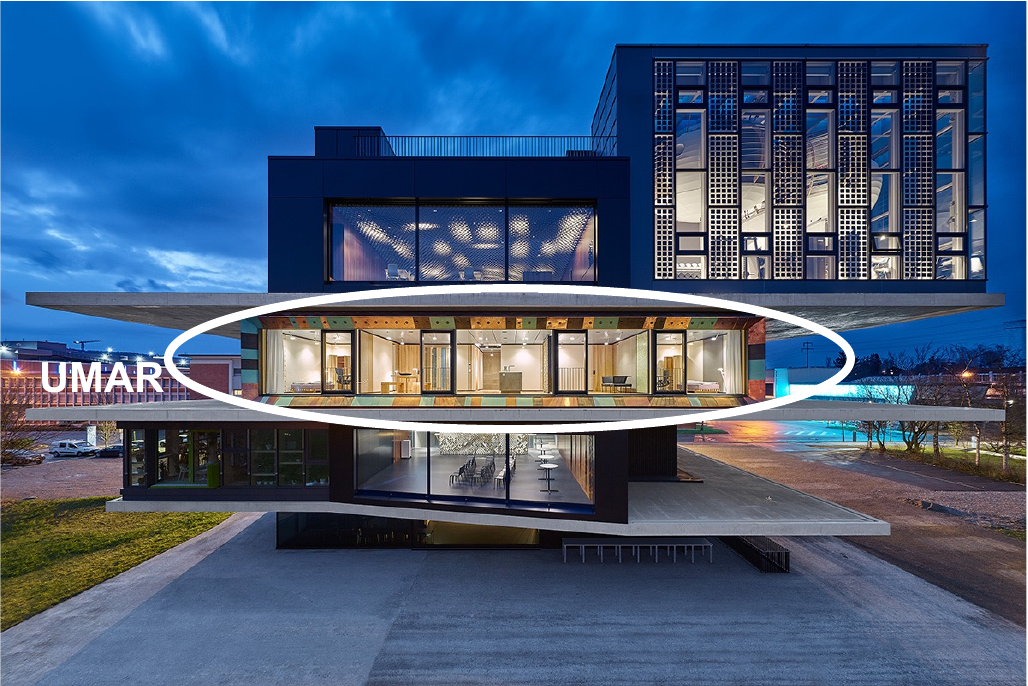}
    \caption{NEST building, Duebendorf, and the UMAR unit circled in white {\copyright} Zooey Braun, Stuttgart.}
    \label{fig: NEST}
    \end{center}
    \end{figure}

In this work, we take advantage of NEST, a vertically integrated district located in Duebendorf, Switzerland, and pictured in Figure~\ref{fig: NEST} \cite{nest}. NEST is composed of several residential and office units, and we focus our attention on \nText{the }"Urban Mining and Recycling" (UMAR) \nText{unit}, where more than three years of data is available to assess the quality of our models.

\remText{\\
Note: Section 4.1 is a new sub-section.}

    \subsection{UMAR}

UMAR is an apartment composed of two bedrooms, with a living room in between them, and two small bathrooms\newText{. We model the temperature of one of the bedrooms throughout this work.}\remText{with effects that are deemed negligible in this work. We model one of the bedrooms throughout this work and only consider the living room as a neighboring zone} All the rooms are equipped with radiant heating/cooling panels in the ceiling and controlled by opening and closing valves to let hot or cold water flow through them depending on the season. Since individual room \newText{power}\remText{energy} consumption measurements are not available, we approximate them by disaggregating the total consumption of UMAR using the design mass flows \remText{of each room} and the amount of time the valves in each room are open. 
\newText{Apart from the temperature and power consumption of the rooms, we also use data about the solar irradiation and the ambient temperature on-site.} \remText{Additional d}\newText{D}etails on the data preprocessing can be found in Appendix~\ref{app: preprocessing}. 

\newText{Since both bathrooms are much smaller and have significantly less heating/cooling power than the bedrooms and the living room, we assume that the heat transfers between the former and the latter are negligible compared to the other heat transfers. In other words, we do not consider the bathrooms as distinct zones and only include the living room as neighboring zone of the modeled bedroom.}

\remText{\\
Note: This Section was moved, it was originally in the methods Section.}

    \subsection{Implementation details} \label{sec: implementation}
    
In our case, 
the inputs $x$ 
are comprised of the raw solar irradiation 
and time information, i.e. the day of the week, the month of the year, and the current time of the day,
and we assume direct control over the power input, i.e. $g(u)=u$. All the models are trained to predict the temperature of the zone over a horizon of three days with time steps of $15$ minutes\newText{, and the data was split in a training and validation set}. \newText{For our experiments, we chose to design $f$ with an encoder-LSTM-decoder structure, where both the encoder and decoder have two layers of $128$ units and the LSTM is composed of two layers of size $512$.} The code of the PCNNs can be found on GitLab\footnote{\url{https://gitlab.nccr-automation.ch/loris.dinatale/pcnn}.} and further \newText{implementation} details \remText{about the implementation} in Appendix~\ref{app: implementation}.

\newText{One}\remText{The} critical implementation point is the initialization of the parameters $a$, $b$, $c$, and $d$ of the PCNN in Figure~\ref{fig: structure}. Indeed, as they are inspired by the known physics of buildings, they must correspond to meaningful values. Furthermore, due to the recurrent use of these parameters to modify the state of the energy accumulator along the prediction horizon, wrong values would have a large impact on the quality of the model and the PCNN might get stuck in a local minimum. In practice, we saw that using \newText{rules}\remText{a rule} of thumb to initialize those parameters to plausible values using our prior knowledge and then letting the PCNN modify them during the back-propagation procedure led to good results, as presented in Section~\ref{sec: accuracy}. We thus use our intuition about how \newText{UMAR}\remText{a building} behaves to define initial values such that:
\begin{itemize}
    \item For $a$ and $d$: The temperature in the zone rises/drops by \SI{1}{\celsius} in \SI{2}{\hour} when \newText{the maximal}\remText{an average} heating/cooling power is applied.
    \item For $b$ and $c$: The temperature drops by \SI{1.5}{\celsius} in \SI{6}{\hour} when the exogenous temperature is \SI{25}{\celsius} lower.
\end{itemize}
\newText{These rules of thumb can be derived from historical data, for example looking at how much time it generally takes for the temperature to rise by \SI{1}{\celsius} when the zone is heated at full power for $a$, respectively to drop by \SI{1.5}{\celsius} when it is \nText{\SI{25}{\celsius}} cold\nText{er} outside and heating is off for $b$. Similar investigations will give plausible initial values for $c$ and $d$.} Since these parameters turn out to be very small in practice, we learn their inverse for numerical stability \newText{(see Appendix~\ref{app: implementation})}.

    \section{Results}
    \label{sec: results}
    
In this section, we analyze the performance of the PCNN that obtained the best error on the validation set of the case study \rText{, using a}\remText{the}\nText{and compare it to a} physics-based RC model \remText{identified as explained in Appendix~\ref{sec: physics_based} }\rText{as} baseline. Since more complex structures \nText{often do not improve the accuracy of RC models, as discussed in Section~\ref{sec: grey-box}}\rText{did not improve the accuracy of the baseline}, \newText{we chose a 2R1C model with two heat sources, the heating/cooling power and the solar gains, \nText{(see Appendix~\ref{sec: physics_based})}\rText{as derived in Appendix~\ref{sec: physics_based}}. This simple architecture furthermore presents the advantage to have a very similar form to the physics-based module $E$ in PCNNs (see Appendix~\ref{app: maths}), which allows us to assess the impact of the NNs in $D$ on the model performance.} 
Note that the RC model has a sampling time of \SI{1}{\minute}, 
we thus keep the \rText{control}\nText{power} input fixed over intervals of \SI{15}{\minute} 
when we compare its predictions with the ones of the PCNN.

    \subsection{\nText{Improving the generalization issue of NNs}}

While we only discuss one PCNN in depth throughout this section, a broader analysis can be found in Appendix~\ref{app: results}, where we trained PCNNs with different random seeds and on the other bedroom in UMAR for comparison. We obtained consistent results, showing the robustness, respectively the flexibility of the approach. 

We additionally performed an ablation study where we removed the physics-inspired prior $E$ and used only the classical encoder-LSTMs-decoder framework \newText{in $f$} to predict the temperature evolution\newText{, concatenating}\remText{ using} all the available features as inputs \rText{in $x$} (hence losing any physical consistency guarantee). Interestingly, as presented in Table~\ref{tab: losses}, while PCNNs could not attain the performance of classical LSTMs on the training data due to their constrained structure to follow the underlying physical laws, they obtained lower errors on the validation set. This confirms that PCNNs solve part of the generalization issue of classical NNs, having a smaller tendency to overfit the training data but retaining enough expressiveness to perform well on new data.

In the rest of this section, \nText{however}, we only investigate in detail the accuracy of the best PCNN compared to the RC model \rText{and not to classical LSTMs} since \rText{the latter} LSTMs were found to be physically inconsistent in our experiments, as pictured in Figure~\ref{fig: inconsistency}\nText{, a critical issue for control-oriented thermal models.} \rText{. They are hence not well-suited to model the bedroom of our case study and we cannot compare them to the proposed PCNN.}


\begin{table}[]
    \centering
    \begin{tabular}{l|c|c|c} \hline
        & Seed & Training loss & Validation loss \\ \hline
        \textit{LSTMs} & $0$ & \remText{$1.31$} \newText{0.57} & \remText{$2.47$} \newText{2.28} \\
        & $1$ & \remText{$1.02$} \newText{0.57} & \remText{$2.56$} \newText{1.92} \\
        & $2$ & \remText{$1.19$} \newText{1.14} & \remText{$2.56$} \newText{2.30} \\ \cline{2-4}
        & \textbf{Mean} & $\textbf{\remText{1.17} \newText{0.76}}$ & $\textbf{\remText{2.53} \newText{2.17}}$ \\
        \hline
        \textit{PCNNs} & $0$ & \remText{$1.80$} \newText{1.83} & \remText{$1.96$} \newText{1.93} \\
        & $1$ & \remText{$2.31$} \newText{1.85} & \remText{$1.63$} \newText{1.65} \\
        & $2$ & \remText{$2.35$} \newText{2.06} & \remText{$1.79$} \newText{1.75} \\ \cline{2-4}
        & \textbf{Mean} & $\textbf{\remText{2.15} \newText{1.91}}$ & $\textbf{\remText{1.79} \newText{1.78}}$ \\
        \hline 
    \end{tabular}
    \caption{Comparison training and validation loss for three classical LSTMs and PCNNs, scaled by $10^{3}$ (full table in Appendix~\ref{app: results}).}
    \label{tab: losses}
\end{table}

    \subsection{Performance analysis}
    \label{sec: accuracy}

\begin{figure}
    \begin{center}
    \includegraphics[width=\columnwidth]{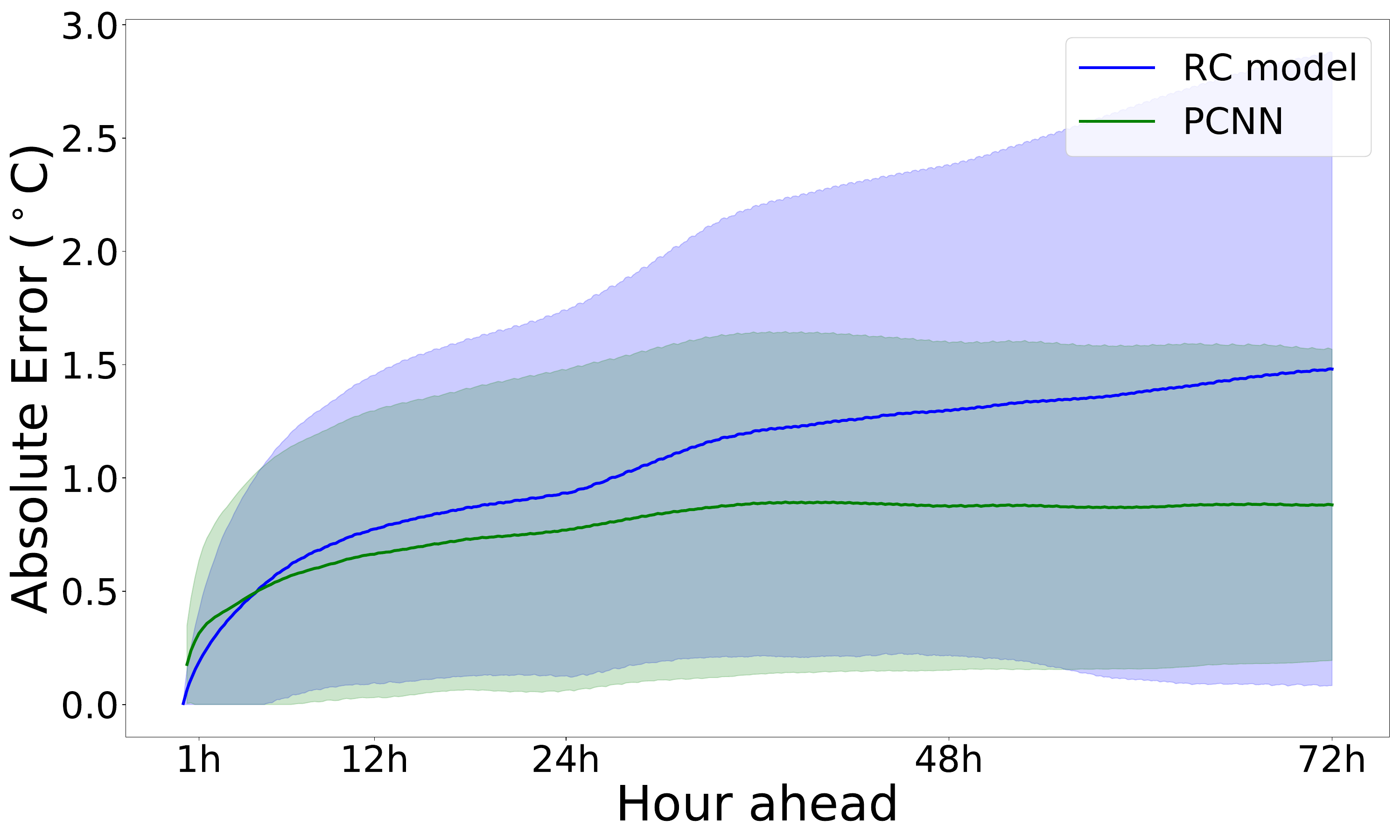}
    \caption{Mean and standard deviation of the error at each time step of the prediction horizon for both the RC model in blue and the PCNN in green, where the statistics were computed from almost $2000$ predictions from the validation set.}
    \label{fig: error propagation}
    \end{center}
    \end{figure}

Since predicting the evolution of the temperature 
for several time steps entails a recursive use of the architecture in Figure~\ref{fig: structure}, 
we leverage the ability of LSTMs to handle long sequences of data to 
minimize the error over the entire horizon. On the other hand, RC models are usually fitted over a single step, 
leading to error propagation, as pictured in Figure~\ref{fig: error propagation}, where we plotted the average \nText{Absolute Error (AE)}\rText{error} and one standard deviation for both models over almost $2000$ possibly overlapping $3$-day long sequences of data from the validation set of the PCNN, i.e. unseen data. Note that while the RC model used as baseline in this work is not optimal, it was nonetheless tuned to obtain good accuracy, with an average error \newText{below}\remText{of} \SI{1}{\celsius} after \SI{24}{\hour}. 

One can observe the PCNN providing better predictions than the RC model in general, which is supported by the \rText{Mean Absolute Errors (MAEs)}\nText{average AE} reported at key points along the horizon in Table~\ref{tab: error propagation}. In particular, the PCNN is able to keep a good accuracy even on long horizons, with an error \newText{more than $40\%$}\remText{almost $50\%$} lower than the RC model after three days. On the other hand, it presents slightly higher errors at the beginning of the horizon because of the warm start that is implemented (Appendix~\ref{app: implementation}): since they firstly predict past data -- the last \SI{3}{\hour} -- PCNNs might indeed start the actual prediction horizon at a temperature different from the true one. 
Nonetheless, since we observed that the warm start benefited the overall performance of PCNNs during our experiments, we kept it in the final implementations.

\begin{table}[]
    \centering
    \begin{tabular}{c|c|c} \hline
        Hours ahead & RC model & \textbf{PCNN (ours)} \\ \hline
        \SI{1}{\hour} & \textbf{\remText{\SI{0.19}{\celsius}} \newText{\SI{0.19}{\celsius}}} & \remText{\SI{0.30}{\celsius}} \newText{\SI{0.31}{\celsius}} \\
        \SI{6}{\hour} & \remText{\SI{0.62}{\celsius}} \newText{\SI{0.58}{\celsius}} & \textbf{\remText{\SI{0.55}{\celsius}} \newText{\SI{0.55}{\celsius}}} \\
        \SI{12}{\hour} & \remText{\SI{0.83}{\celsius}} \newText{\SI{0.78}{\celsius}} & \textbf{\remText{\SI{0.69}{\celsius}} \newText{\SI{0.66}{\celsius}}} \\
        \SI{24}{\hour} & \remText{\SI{1.00}{\celsius}} \newText{\SI{0.93}{\celsius}} & \textbf{\remText{\SI{0.77}{\celsius}} \newText{\SI{0.77}{\celsius}}} \\
        \SI{48}{\hour} & \remText{\SI{1.42}{\celsius}} \newText{\SI{1.30}{\celsius}} & \textbf{\remText{\SI{0.86}{\celsius}} \newText{\SI{0.88}{\celsius}}} \\
        \SI{72}{\hour} & \remText{\SI{1.73}{\celsius}} \newText{\SI{1.48}{\celsius}} & \textbf{\remText{\SI{0.87}{\celsius}} \newText{\SI{0.88}{\celsius}}} \\
        \hline
    \end{tabular}
    \caption{Comparison of the MAE of the two models over the prediction horizon.}
    \label{tab: error propagation}
\end{table}

\begin{figure*}
        \centering
        \begin{subfigure}[t]{\columnwidth}
            \includegraphics[width=\textwidth]{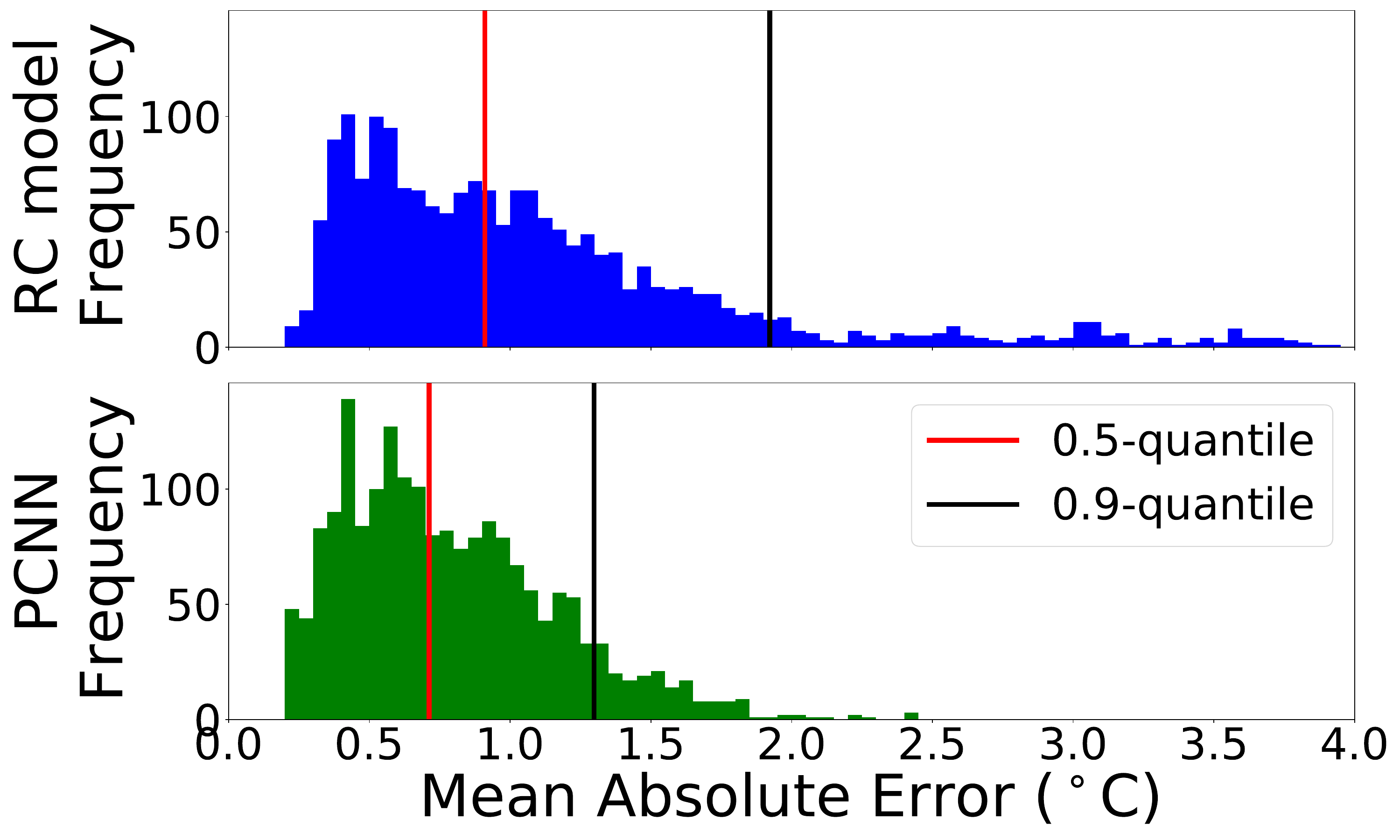}
            \caption{Distribution of the MAE of both models over the test sequences, with the $50\%$ and $90\%$ quantiles marked in red, respectively black.}
        \end{subfigure}
        \hfill
        \begin{subfigure}[t]{\columnwidth}
            \includegraphics[width=\textwidth]{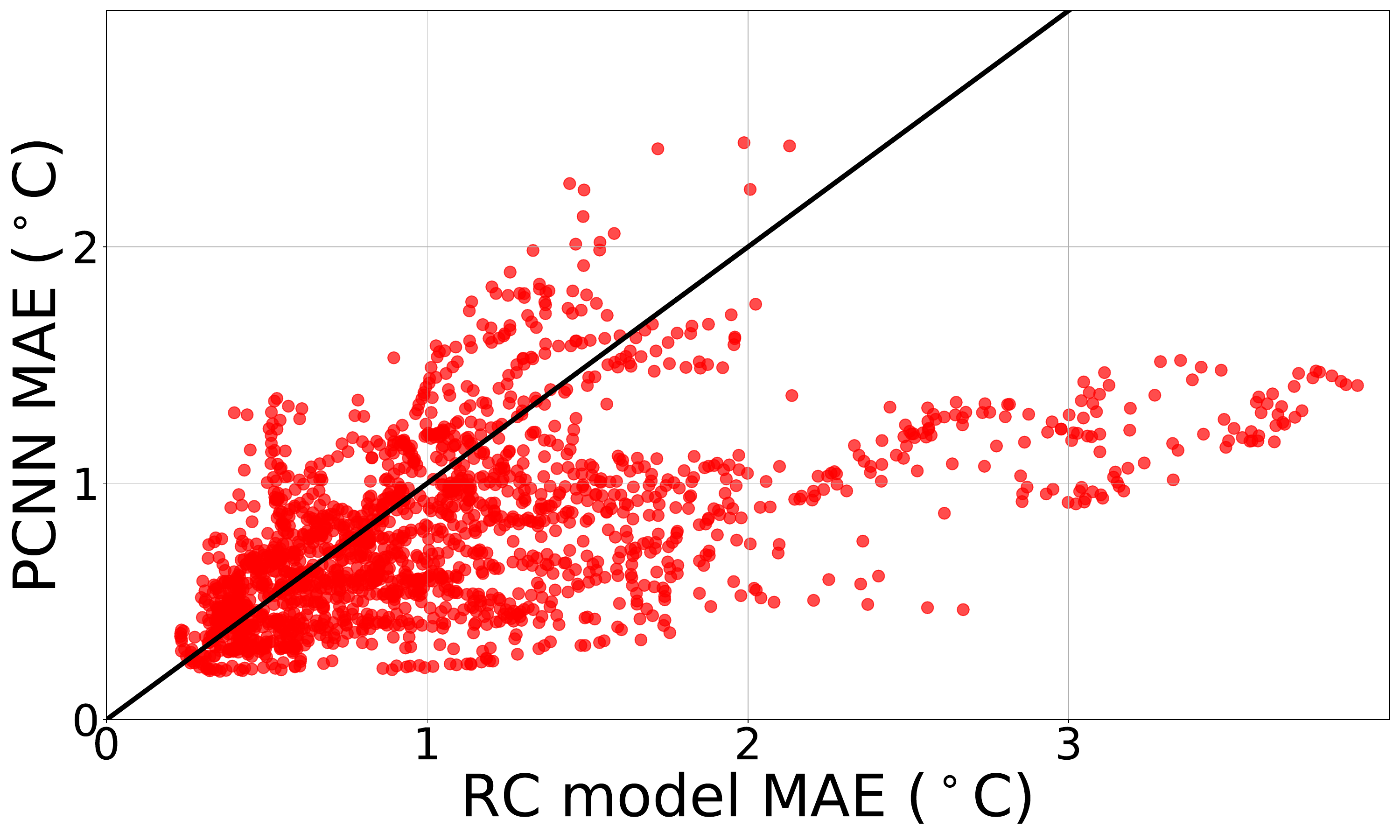}
            \caption{Scatter plot of the MAEs of both models on each test sequence, with the black diagonal line representing equal performance.}
        \end{subfigure}
        \caption{Comparison of the MAE of both the PCNN and RC model over almost $2000$ predictions of three days, taken from the unseen validation data of the PCNN.} 
        \label{fig: error analysis}
    \end{figure*}

\nText{To investigate the Mean Absolute Errors (MAEs) obtained by both models on each sequence of data, we also provide the corresponding histograms and scatter plot in Figure~\ref{fig: error analysis}.}\rText{The histogram and scatter plot of the MAEs obtained by both models on each sequence of data is provided in Figure~\ref{fig: error analysis}.} 
In general, one can see the PCNN dominating the RC model: there are only a few sequences where its error is significantly larger than the one of the RC model, represented by points over the black diagonal line in the right plot. On the other hand, towards the lower and right side of this figure, we find data sequences where the PCNN presents a significantly better accuracy than the RC model. This is confirmed by looking at the error distributions of both models in the left plot, with the errors of the PCNN (green) clustered below \SI{1}{\celsius} and almost always below \SI{2}{\celsius} while the errors of the RC models in blue are much more spread out. This indicates that the PCNN is robust with respect to different inputs, 
even on unseen data. 

Altogether, we can hence conclude that the PCNN is less prone to extreme errors and keeps the majority of errors lower than the given RC baseline, 
proving its robustness and effectiveness. \newText{Remarkably, all the results were obtained on over three years of data, hence under various weather conditions and during all the seasons, which also hints that exogenous variables do not impact the quality of the model much.}


    \subsection{Empirical analysis of the physical consistency}
    \label{sec: empirical}
    
\begin{figure*}
    \begin{center}
    \includegraphics[width=\textwidth]{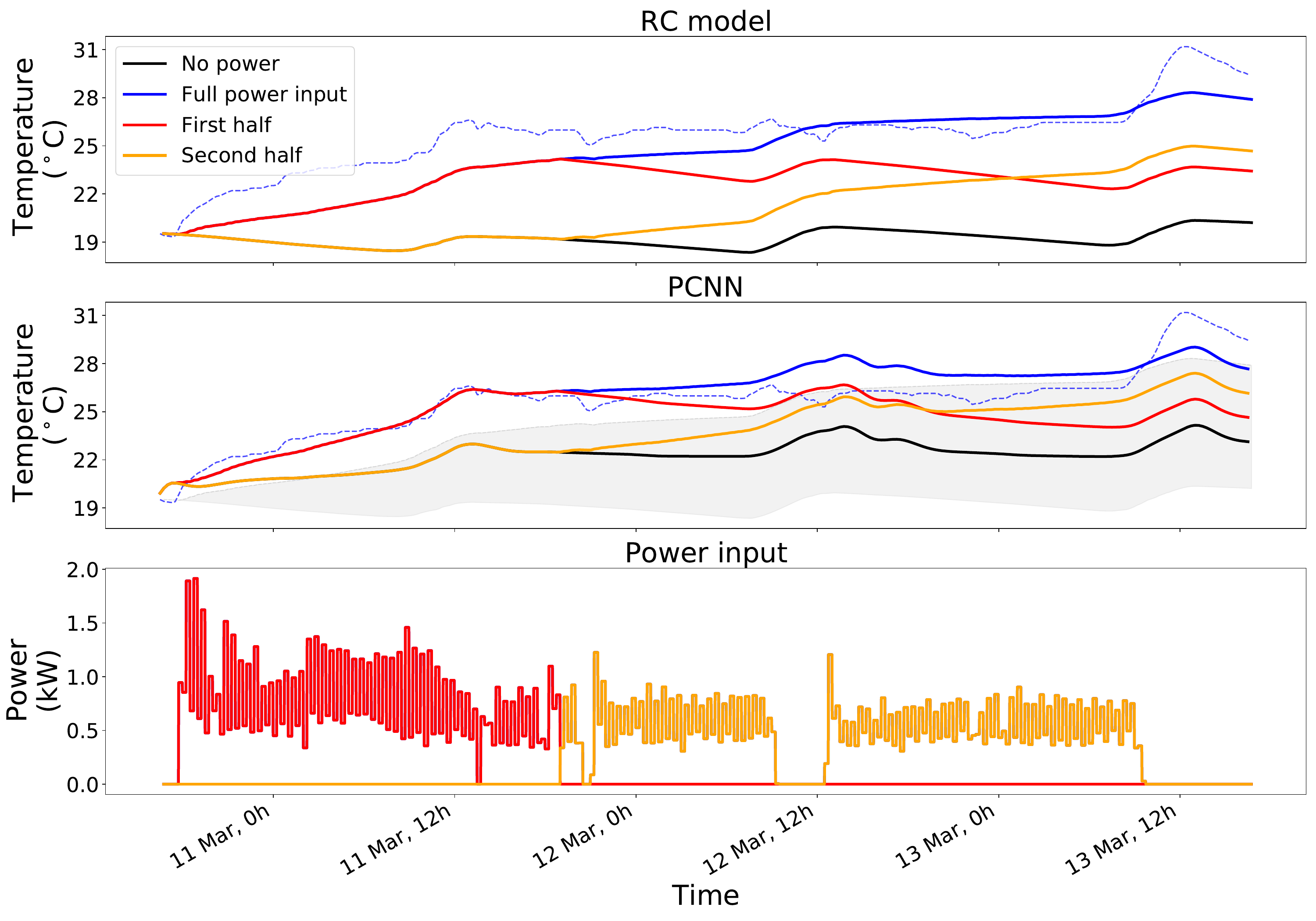}
    \caption{Comparison between the RC model (top) and the PCNN (middle) given the bottom heating control sequence, over three days. In blue, one can assess the precision of both models compared to the ground truth (dashed), where the full control sequence was used. Then, red and orange show the result when only the first half of the control input, respectively the second one, is used. Finally, the black uncontrolled dynamics reflect the case when no power is used, and we shaded the span of the RC model predictions in the middle plot as reference.}
    \label{fig: physical consistency}
    \end{center}
    \end{figure*}

With the physical consistency of the models formally proven in Section~\ref{sec: theory}, we can \nText{now} visualize its impact empirically. Note that the main point of this analysis is to show that the PCNN retains physical consistency even on \textit{unseen} data, i.e. data from the validation set and with various engineered power inputs that do not exist in the data, avoiding the classical generalization issue of NNs detailed in Section~\ref{sec: introduction}. To that end, we take an input sequence from the validation set and compare the temperature predictions of both the RC model and the PCNN when:
\begin{itemize}
    \item the original and true power inputs are applied (blue),
    \item only the first half of the power inputs are used (red),
    \item only the second half of the input is applied (orange),
    \item no power is used (black), hereafter named \textit{uncontrolled},
\end{itemize}
where we separate the power inputs in half with respect to their magnitudes, i.e. so that both the red and orange control sequence apply roughly the same total power. One such experiment is summarized in Figure~\ref{fig: physical consistency} for a heating case, where we also added the ground truth in dashed blue as \rText{a }reference for both models\rText{, to be compared with the full blue lines}. 

Firstly, comparing the blue predictions with the dashed ground truth, we see both models performing well\nText{, exemplifying the results discussed in Section~\ref{sec: accuracy}}. In particular, the proposed PCNN is able to grasp the general trends \rText{well} to match the ground truth despite the large amount of heating power applied and the temperature rising to more than \SI{30}{\celsius}, something unusual in a real setting and hence not well covered by the training data. 

Furthermore, looking at the three other predictions, for which we do not have a ground truth anymore, 
both models again show similar behaviors. This is the visual consequence of the physical consistency proven in Section~\ref{sec: theory}, 
with the red predictions deviating from the blue ones at the same point in time for both models: as soon as we stop heating the room, we get lower temperatures. Similarly, the orange predictions deviate from the uncontrolled dynamics 
at the same points in time for both models. 
Finally, looking at the uncontrolled predictions, one can observe smoother patterns for the PCNNs due to the unforced base dynamics being captured by LSTMs 
instead of the more aggressive linear regression at the core of the RC model. 

\begin{figure}
    \begin{center}
    \includegraphics[width=\columnwidth]{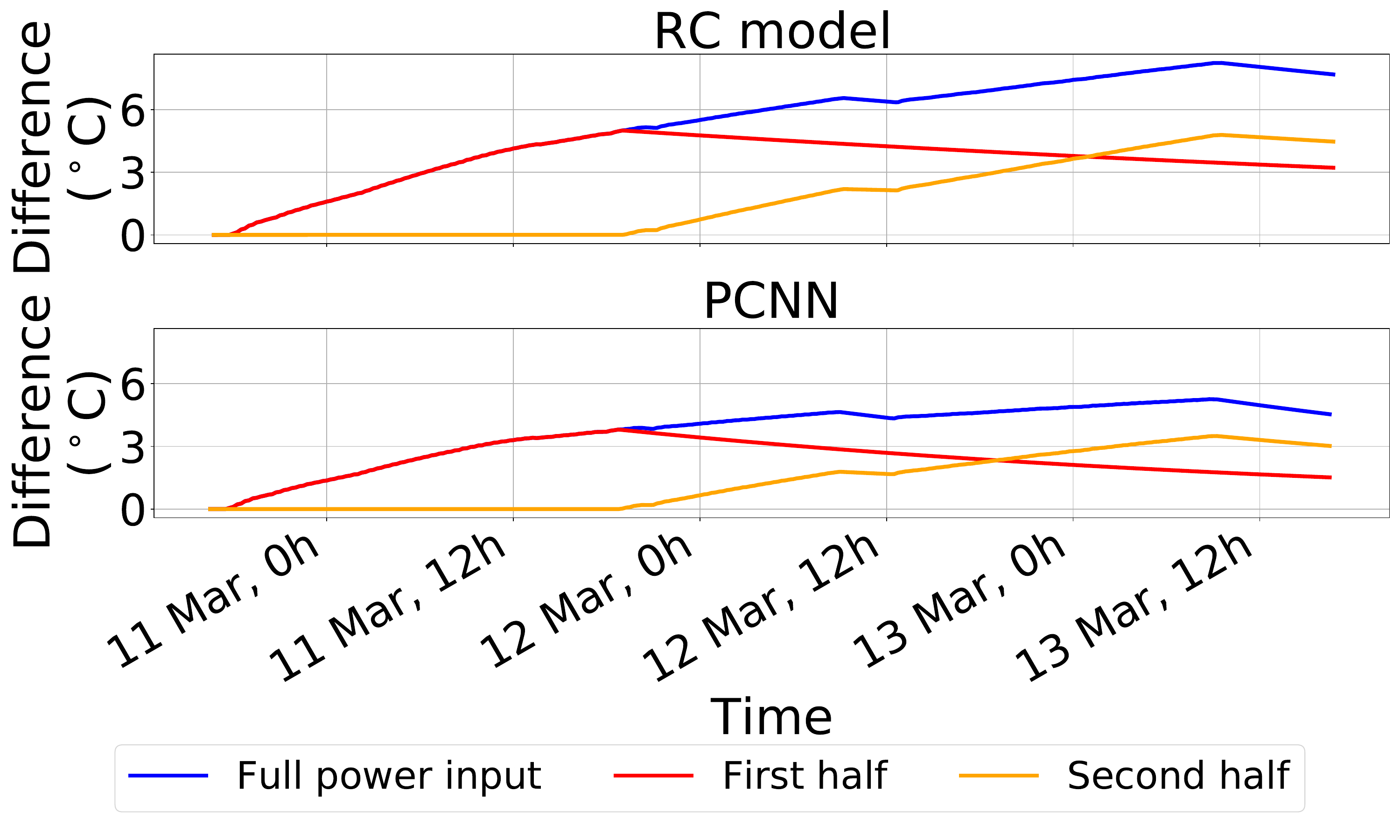}
    \caption{Difference between each control input and the black baseline (no energy) in Figure~\ref{fig: physical consistency}, for the physics-based model (top) and the proposed black-box structure (middle).}
    \label{fig: differences}
    \end{center}
    \end{figure}

To get a better visualization of the behavior of both models with respect to the different control inputs, we can subtract the uncontrolled predictions from the other \rText{ones}\nText{curves}. \rText{This effectively removes the unforced thermal dynamics, which are independent of the power inputs and heat losses to the environment and neighboring zone (Figure~\ref{fig: structure}).} The result is pictured in Figure~\ref{fig: differences} 
and allows us to assess the impact of the three different control sequences on the final predictions. 
As \nText{expected}\rText{one can again directly observe}, both models \nText{still} exhibit similar behaviors, with predictions diverging from the baseline as soon as heating is turned on. On the other hand, when heating is off, the gap with the baseline gets slowly closed because of the higher inside temperature leading to higher energy losses to the environment \nText{and the neighboring zone}. 
Note that the impact of the neighboring room is hard to \rText{remark}\nText{distinguish} in that plot since it is an order of magnitude smaller \nText{than the losses to the outside}. 
    

\rText{One can empirically assess the difference in the parameters $a$, $b$, $c$, and $d$ learned by both models using Figure~\ref{fig: differences}. \rText{We can f}
or example
\rText{see that} $a$ is smaller for the PCNN than for the RC model since the differences with the uncontrolled dynamics are generally smaller, i.e. heating has a smaller impact on the zone temperature. 
Despite not being visible in that plot, we \rText{can}
draw a similar conclusion about the cooling parameter $d$. Concerning heat losses, one can remark that the PCNNs learned parameters $b$ and $c$ entailing roughly the same amount of energy transfer as the RC model, which is particularly observable in the red curve after one day. 
However, we cannot separate the effect of $b$ and $c$ in these plots: the PCNN might have higher heat transfer to the outside than the RC model, but lower heat transfer to the neighboring zone for example. Further analysis would be required to investigate and clarify this relation between $b$ and $c$.}

    \section{Discussion}
    \label{sec: discussion}
    
In this section, we \rText{first} briefly discuss the main differences between PCNNs and classical grey-box models\rText{. We}\nText{ and} then mention potential applications of PCNNs, leveraging their physical consistency, and some hurdles that still need clarification.

    \subsection{Contrasting PCNNs with grey-box models}
    \label{sec: contrast}

Since PCNNs are heavily inspired from classical RC models, we can derive them as a specific form of grey-box models, as detailed in Appendix~\ref{app: contrast}, written as follows:
\begin{align}
    \xi_{k+1} &= \xi_k + m(\xi_k, w^2_k) \nonumber \\
    z_{k+1} &= Az_k + B_ug(u_k) + B_{w^1}w^1_k \label{equ: final RC inspired} \\
            &\qquad + B_d\xi_k + \xi_{k+1} \nonumber
\end{align}
where $z$ represents the state of the system, $u$ the control\nText{lable} inputs, $w^1$ uncontrollable ones, and $\xi$ is a disturbance model computed as a function $m$ of the rest of the uncontrollable inputs $w^2$.
The main structural difference between Equation~(\ref{equ: final RC inspired}) and classical grey-box formulations is the impact of the disturbance $\xi$, which appears both with and without a lag of one in the state update function. 
Traditional approaches generally first forget about the unknown disturbance $\xi$ to identify $A$, $B_u$, and $B_{w^1}$, and then fit a disturbance model to the residuals, e.g. using Gaussian Processes \cite{hewing2019cautious}. 

Despite the similarity with classical grey-box models, PCNNs are fundamentally different both in terms of philosophy and training procedure. Firstly, the linear evolution of the state $z$ captures the main dynamics of grey-box models, including the impact of control inputs, and the nonlinear disturbance $\xi$ corrects them to match the data. On the other hand, in PCNNs, the main (unforced) dynamics $D$ are processed by nonlinear NNs, while the linear energy accumulator $E$ adjusts the predictions according to the controllable inputs and known disturbances, i.e. heat losses. 

Secondly, contrary to classical techniques modeling the disturbance $\xi$ as a separate process, all the parameters of PCNNs are trained simultaneously over the entire prediction horizon -- PCNNs are multi-step-ahead models -- and in an end-to-end fashion to capture dependencies between $D$ and $E$, leveraging automatic BPTT. 

    \subsection{Potential of PCNNs}


As discussed, the good accuracy and physical consistency of PCNNs make them natural candidates for control-oriented zone temperature models. They could however also be used or integrated into Digital Twins (DTs), a fast-growing field that suffers from two problems that PCNNs could solve. Firstly, if historical data is available, they could reduce the effort and knowledge required to calibrate DTs. \nText{Secondly, even when the calibration is successful, DTs often remain slow at run-time due to the level of detail included, something that can be improved by training PCNNs to imitate their behavior and subsequently accelerate the inference procedure.}
\rText{Secondly, once well-calibrated, high-fidelity DTs are often slow at run-time because of the level of detail included in the model, and one could use DTs to generate training data for PCNNs offline and then use the latter for faster inference.} 


PCNNs could also be used in retrofitting operations, albeit restricted to the renewal of energy systems. Indeed, since PCNNs are robust to changes in the control input\nText{s}, one could assess the possibilities arising from different power systems, e.g. the impact of adding or subtracting some heating capacity. Combined with an intelligent controller, one could \nText{for example} anticipate the potential energy savings of the new system and balance them with the installation costs to compute the return on investment of the operation.
 


Overall, we thus see good potential for PCNNs in the field of building modeling and beyond. Indeed, while we present a specific case study on zone temperature models in this work, it is noteworthy that the structure of PCNNs, with a physics-informed module in parallel to a black-box one, is very general and flexible. If more information about the physics of the system is known \newText{or required}, the physics-informed module can be expanded to incorporate it. \newText{For example, if PCNNs are used to analyze the impact of solar gains, then $E$ should include a detailed model of this process on top of the current formulation. Thanks to the modularity of PCNNs, the rest of the pipeline of Figure~\ref{fig: structure} would not be modified, apart from possible changes in which inputs are contained in $x$. Typically, if a detailed model of the solar gains is included in $E$, then the solar irradiation does not need to be processed in $D$ anymore. In general,}\remText{Moreover,} any system with similar underlying physical processes, i.e. systems that can \textit{accumulate} and \textit{dissipate} energy, might be modeled with PCNNs, adjusting the physics-informed module.

    \subsection{Limits of our application}

Firstly, it is important to note that the proposed structure does not fully solve the generalization issue of NNs: PCNNs are only physically consistent with respect to control inputs and exogenous temperatures, i.e. they satisfy the conditions in Equations~(\ref{equ: consistency power})-(\ref{equ: consistency neigh}). Should other inputs in $x$ vary, we cannot guarantee the robustness of the model anymore. In particular, the current version of PCNNs does not meet the condition 
in Equation~(\ref{equ: consistency sun}). This is left for future work for two main reasons: one usually does not have direct access to solar gains through the windows of the zone to model, but only to the horizontal irradiation measurement, which has a nonlinear effect on the zone temperature. 
Furthermore, Equation~(\ref{equ: consistency sun}) is not strictly needed in control-oriented models, the main target of this work. Indeed, modifying the heating or cooling power inside a zone only impacts its temperature, and hence heat transfers indirectly, but the solar gains always have the same impact under any control input.

Secondly, it would be of interest to investigate the tuning of the various parameters, both to optimize the black-box $f$ and, in particular, to understand how to initialize and learn meaningful values for $a$, $b$, $c$, and $d$. 
In this work, we initialized them using prior knowledge and \newText{rules}\remText{a rule} of thumb, 
it remains unclear how and why they get to their final values, which can differ depending on the random seed used during training. Furthermore, we observed that the quality of the solution can vary significantly if they are initialized to unrealistic values, i.e. PCNNs do not always recover physically consistent parameters from data. 
Consequently, it might be useful to add constraints to ensure they retain meaningful values throughout training, i.e. they continuously meet the conditions in Equation~(\ref{equ: conditions}). In practice, however, we did not have that issue, with the parameters only slowly changing around their physics-informed initial value and thus staying physically consistent throughout our experiments.

Lastly, one should keep in mind that the RC baseline used in this paper, albeit tuned to obtain satisfying performance, still remains a low complexity model. While our PCNN was able to get better accuracy than this RC model, it might be possible to find better \nText{physics-based }models. Nonetheless, PCNNs seem very competitive and avoid any need for engineering, which makes them attractive in general.


    \section{Conclusion}
    \label{sec: conclusion}
    

In this work, we presented a novel neural network architecture, dubbed PCNN, with an application to building zone temperature modeling. By treating some input variables separately from the main NN in a physics-informed module, PCNNs include prior knowledge in their structure. They are hence able to capture parts of the underlying physics 
while leveraging the accuracy of NNs to attain significant performance improvement over classical physics-based models without any engineering overhead.

A key \rText{aspect}\nText{advantage} of PCNNs \nText{over existing NN-based thermal models }is that \nText{we could formally prove that their temperature predictions remain}\rText{they are formally proven to be} physically consistent with respect to \nText{any} control inputs and exogenous temperatures \nText{and} over the entire prediction horizon.
\nText{Furthermore, grounding PCNNs in the underlying physics allowed us to mitigate the usual generalization issue of classical NN frameworks.}
\rText{Thus, they represent a credible option to counter the usual generalization issue of NNs when modeling physical systems.}

\rText{This was}\nText{These results were} confirmed by our experiments on \nText{a bedroom temperature modeling case study}\rText{the case study bedroom}, \rText{which showed that PCNNs generally obtain}\nText{in which PCNNs generally obtained} better results than classical LSTMs over the validation data, even when they \rText{are}\nText{were} performing worse during the training phase, strongly indicating that they do not suffer from generalization issues as much as classical frameworks. \nText{Additionally, PCNNs clearly outperformed the RC model baseline while following the underlying physical laws, reducing the error by more than $40\%$ at the end of the $3$-day long prediction horizon.}



Since PCNNs are solely based on data and do not require any engineering, they are very flexible and easy to use. This makes them interesting for different applications in the field of building modeling and beyond. To complete the analysis of their potential, it would be of interest to assess the sensitivity of PCNNs with respect to the key parameters of the physics-informed module, which should have links with physical quantities, and the amount of data required to attain satisfactory performance. In future work, we plan to focus our research on extending the current architecture to the multi-zone setting and use PCNNs in various control schemes to learn intelligent temperature controllers.

    \section*{Acknowledgements}
    
This research was supported by the Swiss National Science Foundation under NCCR Automation, grant agreement 51NF40\_180545.

    \section*{Declaration of competing interests}
    
The authors declare that they have no known competing financial interests or personal relationships that could have appeared to influence the work reported in this paper.

\printcredits

\bibliographystyle{els-cas-templates/model1-num-names}

\bibliography{biblio.bib}

\appendix
    \section*{Appendices}

    \section{RC building model}
    \label{sec: physics_based}
    
    \subsection{General RC models}
    \label{app: general rc}
    
In general, we can describe the thermal dynamics of a room with the following ordinary differential equation (ODE):
\begin{align}
    C\frac{dT}{dt}  &= \frac{dQ^{heat}}{dt} + \frac{dQ^{irr}}{dt} + \frac{dQ^{occ}}{dt} \nonumber \\
                    &\qquad + \sum{\frac{dQ^{cond}}{dt}} + \sum{\frac{dQ^{conv}}{dt}},
\end{align}
where $T$ is the temperature, $C$ the heat capacitance of the air mass, $Q$ respectively represents heat flows from the heating/cooling system (negative values represent cooling energy), the solar irradiation, the occupants, heat conduction, and heat convection, respectively, where both sums are taken over the number of surfaces adjacent to the measured volume of air. 

In this work, we consider conductive and convective transfer together in two heat transfers: one to represent transfer to the neighboring zone (assuming there is only one) and the other to gather losses to the environment, both being proportional to the corresponding temperature gradient. 
Additionally, we process the horizontal solar irradiation data 
to reflect the solar gains through the windows as follows:
\begin{align}
Q^{irr}(t) &= \frac{\sin{(\theta - \theta_0)}\cos{(\phi)}}{\sin{(\phi)}}I(t), \label{equ: irradiation}
\end{align}
where $I$ is the irradiation measured by the sensor, $\phi$ the altitude and $\theta$ the azimuth of the sun, and $\theta_0$ accounts for the orientation of the window (as counter clock-wise rotation from a north-south-aligned surface facing east).
Altogether, we can then rewrite the thermal dynamics as:
\begin{align}
    \frac{dT}{dt}   &= \frac{1}{C} \frac{dQ^{heat}}{dt} + \frac{\epsilon}{C}\frac{dQ^{irr}}{dt} + \frac{\eta}{C}\frac{dQ^{rest}}{dt} \nonumber \\
                    &\qquad - \frac{1}{C R_{out}}\frac{d(T - T^{out})}{dt} \label{equ: no need} \\
                    &\qquad - \frac{1}{C R_{neigh}}\frac{d(T - T^{neigh})}{dt} \nonumber
\end{align}
with $\epsilon$ representing the lumped permissivity of the windows and exterior walls, $R_{out}$ and $R_{neigh}$ the thermal resistance of the walls adjacent to the outside, respectively the neighboring zone, and $T^{out}$ and $T^{neigh}$ the temperature outside, respectively in the neighboring zone. We then group all the other heat gains in $Q^{rest}$, scaled by a parameter $\eta$ and discretize this ODE with the Euler forward method and the time step $\Delta_t=\SI{1}{\minute}$, yielding:
\begin{align}
    T_{k+1} &= T_k + \Delta_t * [\frac{1}{C} Q^{heat}_k + \frac{\epsilon}{C}Q^{irr}_k + \frac{\eta}{C}Q^{rest}_k \nonumber \\
            &\quad \qquad - \frac{1}{C R_{out}}(T_k - T^{out}_k) \label{equ: no need 2} \\
            &\quad \qquad - \frac{1}{C R_{neigh}}(T_k - T^{neigh}_k)] \nonumber 
\end{align}
Grouping the constants together and defining new parameters $a$, $b$, $c$, $e_1$ and $e_2$, we can reformulate it as follows:
\begin{align}
    T_{k+1} &= T_{k} + a Q^{heat}_k - b(T_k - T^{out}_k) \nonumber \\
            &\quad - c(T_k - T^{neigh}_k) + e_1 Q^{irr}_k + e_2 Q^{rest}_k \label{equ: physics-based 2}
\end{align}

    \subsection{Baseline RC model}
    \label{app: baseline}
    
In this work, to create a simple RC model to use as a comparison baseline, we assume no knowledge of the occupants and other heat gains and discard the corresponding term $e_2 Q^{rest}_k$. Rewriting Equation~(\ref{equ: physics-based 2}), we get:
\begin{align}
T_{k+1} - T_{k} &= \begin{bmatrix} Q^{heat}_k \\ - (T_k - T^{out}_k) \\ - (T_k - T^{neigh}_k) \\ Q^{irr} \end{bmatrix} ^T \begin{bmatrix} a \\ b \\ c \\ e_1 \end{bmatrix}  \nonumber \\
\Delta T_{k+1} &= y_k^T p, 
\end{align}
where $\Delta T$ represents the temperature difference, $y$ groups the factors influencing it, and $p$ the unknown parameters. Doing this for every time step, we can create matrices of data, grouping all the temperature differences in matrix $X$ and the external factors in $Y$:
\begin{align}
\begin{bmatrix} \Delta T_1 \\ \vdots \\ \Delta T_N \end{bmatrix} &= \begin{bmatrix} y_1^T \\ \vdots \\ y_N^T \end{bmatrix} p \nonumber \\
X &= Yp
\end{align}
Finally, we can use Least Squares to identify the parameters:
\begin{align}
Y^T X &= Y^T Y p \nonumber \\
p &= (Y^T Y)^{-1} Y^T X
\end{align}

    \section{Mathematical derivations}
    \label{app: maths}
    
    \subsection{Physics-based predictions}
    \label{proof: physics-based}
    
We can rewrite the predictions of the RC model from Equation~(\ref{equ: physics-based 2}) as follows:
\begin{align}
    T_{k+1} &= (1-b-c)T_{k} + a Q^{heat}_k + b T^{out}_k \nonumber \\
            &\quad + c T^{neigh}_k + e Q^{irr}_k,
\end{align}
Applying this transformation recursively yields the following two-steps-ahead temperature predictions:
\begin{align}
    T_{k+2} &= (1-b-c)T_{k+1} + a Q^{heat}_{k+1} + b T^{out}_{k+1} \nonumber \\
            &\quad+ c T^{neigh}_{k+1} + e Q^{irr}_{k+1} \nonumber \\
            &= (1-b-c)[(1-b-c)T_{k} + a Q^{heat}_{k} \nonumber \\
            &\quad \qquad + b T^{out}_k + c T^{neigh}_k + e Q^{irr}_k] \nonumber \\
            &\quad+ a Q^{heat}_{k+1} + b T^{out}_{k+1} + c T^{neigh}_{k+1} + e Q^{irr}_{k+1} \\
            &= (1-b-c)^2T_{k}  \nonumber \\
            &\quad + (1-b-c)[a Q^{heat}_{k} + b T^{out}_k\nonumber  \\
            &\quad \qquad + c T^{neigh}_k + e Q^{irr}_k] \nonumber \\
            &\quad + a Q^{heat}_{k+1} + b T^{out}_{k+1} + c T^{neigh}_{k+1} + e Q^{irr}_{k+1} \nonumber
\end{align}
This leads to the general formula below for the temperature prediction $i$ time steps ahead:
\begin{align}
    T_{k+i} &= (1-b-c)^i T_{k} \nonumber \\ 
            &\quad + \sum_{j=1}^{i}(1-b-c)^{(j-1)} {[a Q^{heat}_{k+i-j}} \\ 
            &\quad \qquad {+ b T^{out}_{k+i-j} + c T^{neigh}_{k+i-j} + e Q^{irr}_{k+i-j}]} \nonumber \label{equ: final rc}
\end{align}
Remarkably, this model is known to follow the laws of physics by design, i.e. it satisfies Equations~(\ref{equ: consistency power})-(\ref{equ: consistency sun}) as long as all the parameters $a$, $b$, $c$, and $e$ are \newText{small and} positive. This is true for real systems since they represent small positive physical constants, i.e. inverses of resistances and capacitances.

    \subsection{Black-box predictions}
    \label{proof: black-box}

PCNN predictions from Equation~(\ref{equ: PCNN}) can be rewritten as follows:
\begin{align}
    T_{k+1} &= (1-b-c) T_k + f(D_k, x_k) + a g(u_k) \nonumber \\
            &\quad + b T^{out}_k + c T^{neigh}_k 
\end{align}
When this formula is applied recursively, what the model does in practice, we get:
\begin{align}
    T_{k+2} &= (1-b-c) T_{k+1} + f(D_{k+1}, x_{k+1}) \nonumber \\
            &\quad + a g(u_{k+1}) + b T^{out}_{k+1} + c T^{neigh}_{k+1} \nonumber \\
            &= (1-b-c) [(1-b-c) T_k + f(D_k, x_k) \nonumber \\
            &\quad \qquad+ a g(u_k) + b T^{out}_k + c T^{neigh}_k] \\
            &\quad + f(D_{k+1}, x_{k+1}) + a g(u_{k+1}) \nonumber \\
            &\quad + b T^{out}_{k+1} + c T^{neigh}_{k+1} \nonumber
\end{align}
Note that $D_{k+1} = D_k + f(D_k, x_k)$ is independent from all the other variables $u$, $T^{out}$, and $T^{neigh}$. We can thus keep the recursion as is, only noting that at each time step $D$ goes through the neural network function $f$ so that we end up with a nested application of $f$ to the inputs $x$. However and crucially, they do not get impacted by changes of control input, and we can write temperature predictions from the model as follows:
\begin{align}
    T_{k+2} &= (1-b-c)^2 T_k + (1-b-c)[f(D_k, x_k) \nonumber \\
            &\quad \qquad+ a g(u_k) + b T^{out}_k + c T^{neigh}_k] \nonumber \\
            &\quad + f(D_{k+1}, x_{k+1}) + a g(u_{k+1}) \\
            &\quad + b T^{out}_{k+1} + c T^{neigh}_{k+1} \nonumber
\end{align}
Similarly to the physics-based case, this leads to the following general formula for any future predictions:
\begin{align}
    T_{k+i} &= (1-b-c)^i T_k \nonumber \\
            &\quad + \sum_{j=1}^{i}(1-b-c)^{(j-1)} {[f(D_{k+i-j}, x_{k+i-j})}  \\
            &\quad \qquad + a g(u_{k+i-j}) + b T^{out}_{k+i-j} + c T^{neigh}_{k+i-j}] \nonumber
\end{align}

    \section{Data preprocessing}
    \label{app: preprocessing}
    
    \subsection{NEST data}
    
Data from all the sensors in NEST is sampled and stored at a frequency of one minute. 
Concerning the solar irradiation data, we delete constant streaks of more than \SI{20}{\hour} than indicate a fault of the sensor -- where \textit{deleting} refers to setting the values to \textit{NaN} -- and clip the measurement at $0$ since it cannot be negative. For the outside temperature, we delete constant streaks of more than \SI{30}{\minute}. Both measurements are then smoothed with a Gaussian filter with $\sigma=2$. For power inputs, we delete constant streaks of more than $1$ day and smooth the measurements with a Gaussian filter with $\sigma=1$. Finally, the temperature measurements in both the room of interest and the neighboring one are smoothed with $\sigma=5$.

Before using the created data, we linearly interpolate all the missing values when less than \SI{30}{\minute} of information is missing. When we use it to train and test PCNNs, the data is subsampled to $15$ minute intervals through averaging.

    \subsection{Individual room energy consumption}
    \label{app: designmassflow}
    
As mentioned in Section~\ref{sec: case study}, UMAR has a unique power meter and we need to disaggregate this global measurement $P^{tot}$ into individual consumption for each room. To that end, we use the design mass flow $\dot m^i$ of room $i$, something known from technical construction sheets. At each time step $t$, we then approximate the power consumed by each room, $P^i$, as follows:
\begin{align}
    P^i_t = \frac{u^i_t \dot m^i_t}{\sum_{k}{u^k_t \dot m^k_t}} P^{tot}_t,
\end{align}
where $u^i$ is the amount of time the valves are opened and we sum over all the $k=5$ rooms in UMAR. In words, we approximate the individual energy consumption of each to be proportional to the amount of water flowing through its ceiling panels.

    \section{Implementation details}
    \label{app: implementation}

The month and time of day variables are represented by sine and cosine functions to introduce periodicity, so that the last month has a value close to the first month of the year for example. Mathematically, two variables are created:
\begin{align}
    t^{sin}_m &= \sin{(\frac{m}{12} 2 \pi)}, &t^{cos}_m &= \cos{(\frac{m}{12} 2 \pi)}, \label{equ: time}
\end{align}
where the months $m$ are labeled linearly and in order from $1$ to $12$. The same processing is done for the time step in during the day, replacing the factor $12$ in Equation~(\ref{equ: time}) by $96$, the number of \SI{15}{\minute} interval in one day.

\remText{For our experiments, we chose to design $f$ with an encoder-LSTM-decoder structure, where both the encoder and decoder have two layers of $128$ units, and the LSTM is composed of two layers of size $512$.} We let the initial hidden and cell state of the LSTM be learned during training and additionally give the model a warm start of \SI{3}{\hour}, i.e. the PCNN first predicts the last $12$ time steps in the past, where we feed the true temperatures back to the network to initialize all the internal states, before predicting the temperature over the given horizon. We train the PCNN to minimize the Mean Square Error (MSE) of the predictions over a horizon of $3$ days with $15$-minute time steps, and use the Adam optimizer with a decreasing learning rate of $\frac{0.001}{\sqrt{h}}$ at epoch $h$. We create sequences of data using sliding windows of minimum \SI{12}{\hour} -- and maximum $3$ days -- with a stride of \SI{1}{\hour}. We then separate both the heating and cooling season in training and validation data with an $80$-$20\%$ split to ensure a fair partition of heating and cooling cases in the training and validation sets. Finally, the data is normalized between $0.1$ and $0.9$.

Since $a$, $b$, $c$, and $d$ are very small values that could be unstable during training -- hence leading to physically inconsistent parameters --, we rewrite:
\begin{align}
    s &= s_0 \tilde s, \quad \forall s \in \{a,b,c,d\} \nonumber
\end{align}
where $s_0$ is the initial value of the parameter. We initialize $\tilde s = 1$ and let the backpropagation algorithm modify this much more stable value instead.


    \section{Additional results}
    \label{app: results}
    
    \subsection{Robustness of PCNNs}
    
\begin{figure}
    \begin{center}
    \includegraphics[width=\columnwidth]{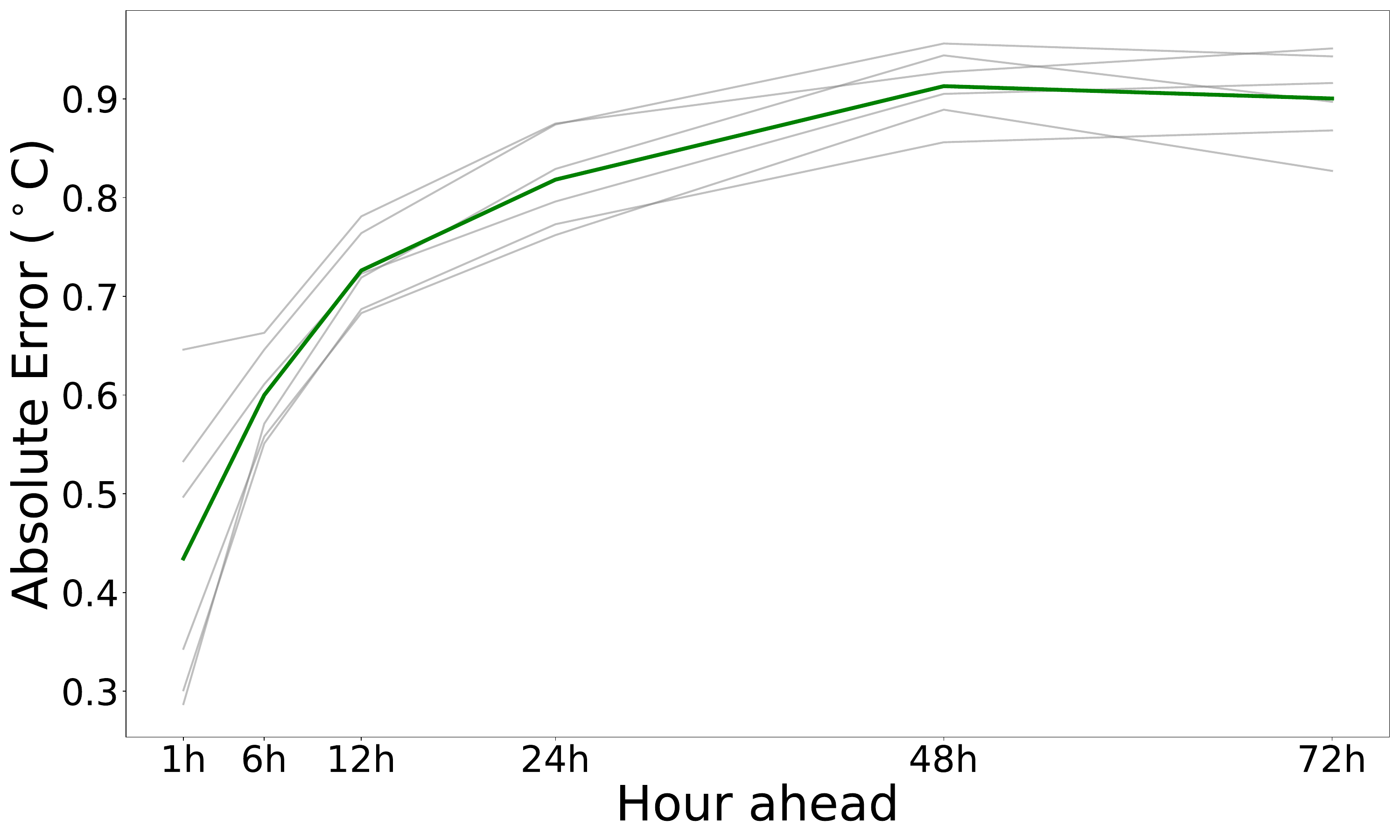}
    \caption{MAE of six PCNNs with different random seeds at six chosen prediction steps in grey and the average in green, where the statistics were computed from almost $2000$ predictions from the validation set.}
    \label{fig: comparison paper}
    \end{center}
    \end{figure}

To further analyze the robustness of the PCNN discussed in Section~\ref{sec: results}, we trained five other networks with the same structure but different random seeds. As pictured in Figure~\ref{fig: comparison paper}, all six models provide similar accuracy over the validation set, except at the beginning of the horizon. Two out of the six PCNNs trained indeed showed oscillatory behavior on the first prediction steps, leading to higher errors.
    
    \subsection{Learned parameters}
    
To complete the analysis of the PCNN presented in Section~\ref{sec: results}, we also display the final values of the parameters $a$, $b$, $c$, and $d$ in Table~\ref{tab: final parameters}. 
Overall, we see that the parameters do not change much, and the same conclusion was drawn for the other PCNNs trained during our experiments. Out of the six PCNNs plotted in Figure~\ref{fig: comparison paper}, only two modified the values substantially, even though by a maximum of $10\%-15\%$, and they correspond to the two models showing the worst performance overall. 

\begin{table}[]
    \centering
    \begin{tabular}{c|c|c} \hline
         & Starting & \textbf{Learned} \\
        Parameter & value & \textbf{value} \\ \hline
        $a$ & $2$ & \remText{$1.98$} \newText{$2.01$} \\
        $b$ & $1.5$ & \remText{$1.50$} \newText{$1.50$} \\
        $c$ & $1.5$ & \remText{$1.51$} \newText{$1.51$} \\
        $d$ & $2$ & \remText{$2.00$} \newText{$1.97$} \\\hline
    \end{tabular}
    \caption{Comparison between the initial and learned values of the PCNN parameters, in degrees Celsius. For $a$ and $d$, it represents how many degrees are gained in \SI{4}{\hour} \newText{when heating/cooling at full power}\remText{with an average heating/cooling power}, while for $b$ and $c$ it represents how many degrees are lost through heat transfer in \SI{6}{\hour} when the exogenous temperature is \SI{25}{\celsius} lower.}
    \label{tab: final parameters}
\end{table}

    \subsection{Flexibility of PCNNs}
    
\begin{figure}
    \begin{center}
    \includegraphics[width=\columnwidth]{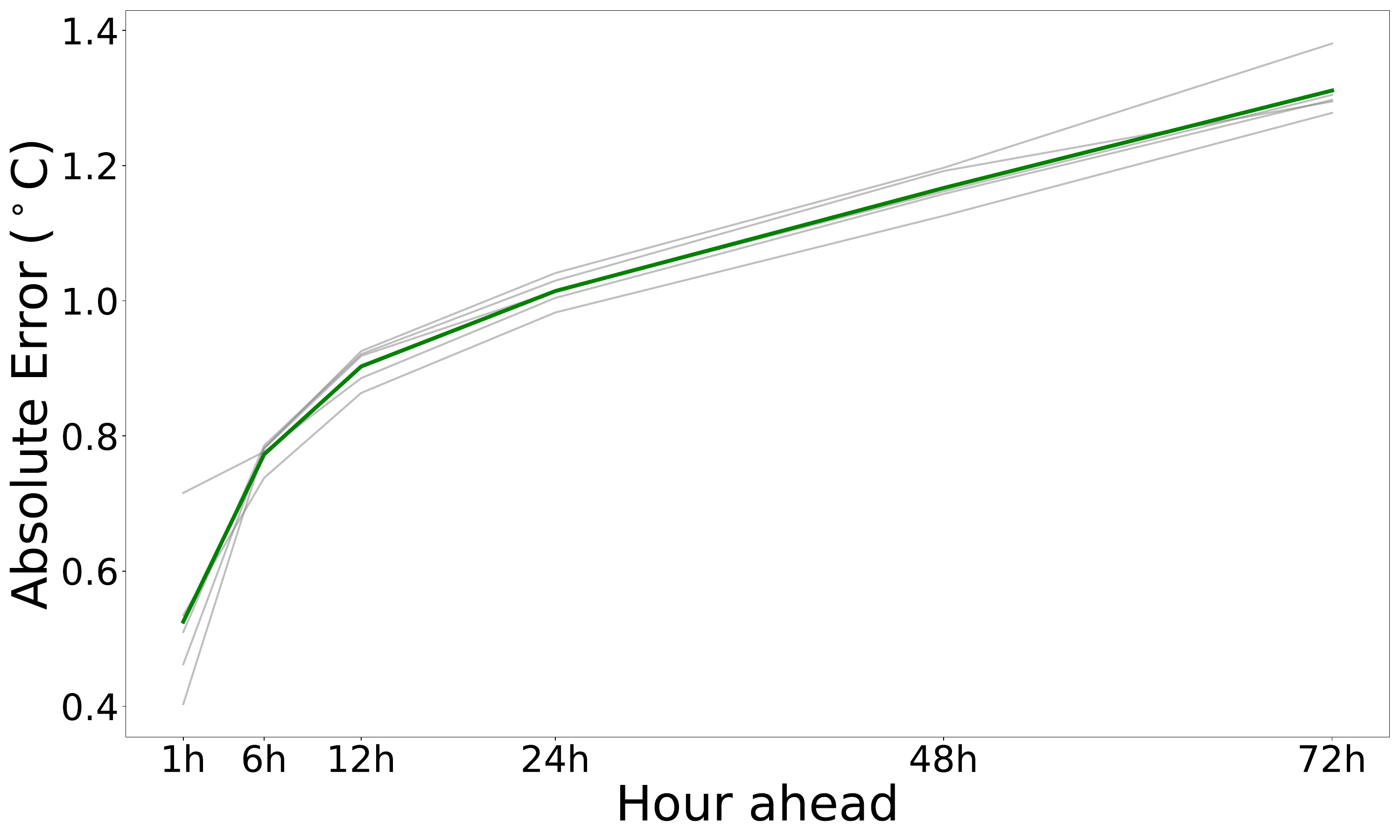}
    \caption{MAE on the other bedroom in UMAR at key time steps of the prediction horizon for the PCNN with five different random seeds, where the statistics were computed from almost $2000$ predictions from the validation set.}
    \label{fig: comparison 272}
    \end{center}
    \end{figure}

Additionally, to test the flexibility of our approach, we trained five PCNNs on the other bedroom in UMAR, again with five different random seeds. As can be observed in Figure~\ref{fig: comparison 272}, the models again arrive at a similar accuracy to what was obtained for the first bedroom (\newText{see} Figure~\ref{fig: comparison paper}), except \newText{towards the}\remText{at the very} end of the prediction horizon, where the error is \newText{$20-40\%$}\remText{around $20\%$} higher. Nonetheless, the performance of PCNNs is comparable for both rooms, which is particularly interesting since no engineering was required to transfer the model between them: we used the same architecture for both bedrooms, simply changing the training and validation data sets. The training and validation errors displayed in Table~\ref{tab: app losses 272} confirm these conclusions.
    
    \begin{table}[]
    \centering
    \begin{tabular}{c|c|c} \hline
        Seed & Training loss & Validation loss \\ \hline
        $0$ & \remText{$1.39$} \newText{$1.82$} & \remText{$2.49$} \newText{$2.42$} \\
        $1$ & \remText{$1.97$} \newText{$1.66$} & \remText{$2.61$} \newText{$2.44$} \\
        $2$ & \remText{$1.71$} \newText{$1.58$} & \remText{$2.55$} \newText{$2.52$} \\
        $3$ & \remText{$1.81$} \newText{$1.66$} & \remText{$2.57$} \newText{$2.54$} \\
        $4$ & \remText{$1.82$} \newText{$1.66$} & \remText{$2.50$} \newText{$2.39$} \\ \hline
        \textbf{Mean} & \remText{$\textbf{1.74}$} \newText{$\textbf{1.68}$} & \remText{$\textbf{2.54}$} \newText{$\textbf{2.46}$} \\
        \hline
    \end{tabular}
    \caption{Training and validation losses of five PCNNs on the other bedroom in UMAR, scaled by $10^{3}$.}
    \label{tab: app losses 272}
    \end{table}
    
    \subsection{Comparison to classical LSTMs}
    
\begin{figure}
    \begin{center}
    \includegraphics[width=\columnwidth]{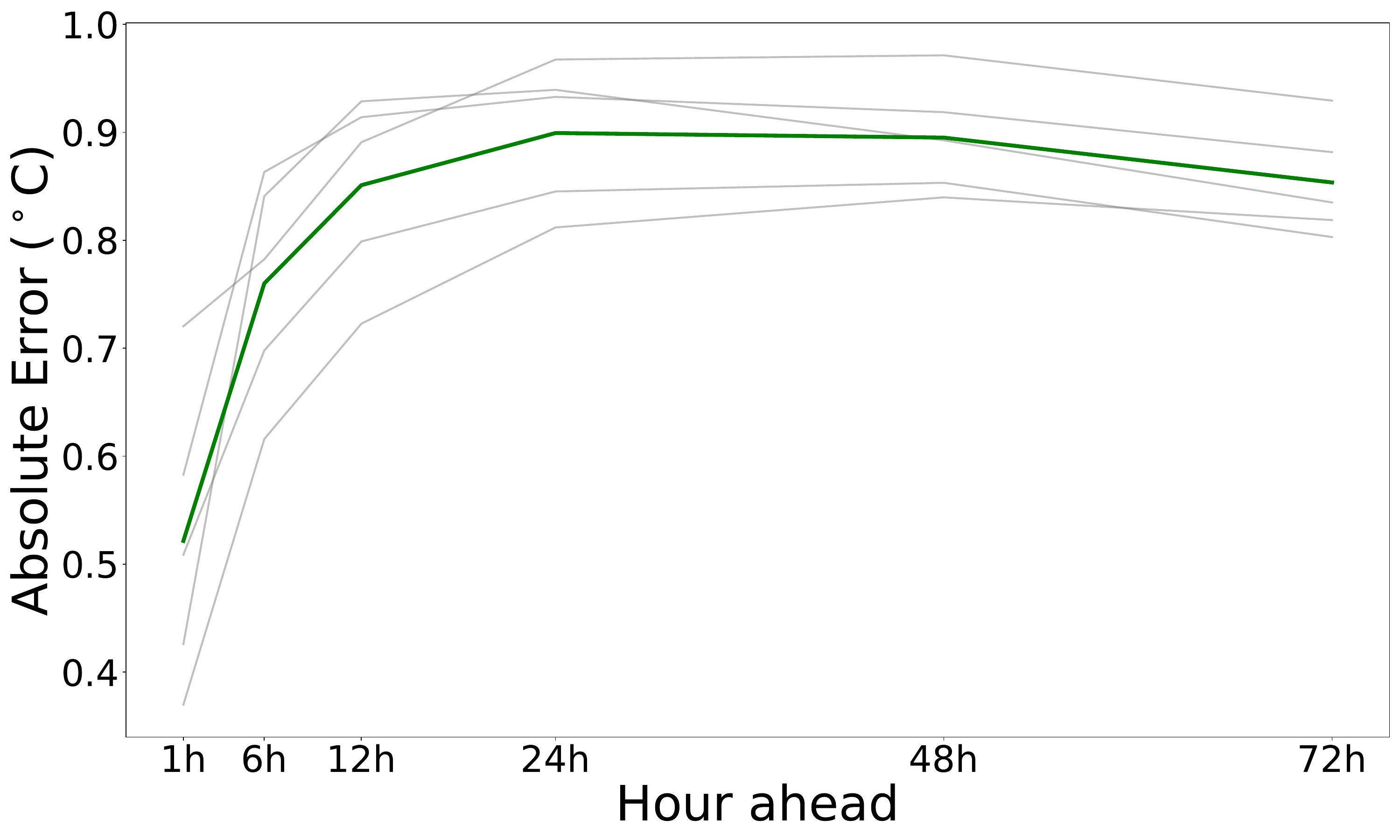}
    \caption{MAE at key time steps of the prediction horizon for the classical encoder-LSTMs-decoder network with five different random seeds in grey and the average in green, where the statistics were computed from almost $2000$ predictions from the validation set.}
    \label{fig: comparison LSTMs}
    \end{center}
    \end{figure}

Finally, to analyze the impact of the prior knowledge inclusion in $E$, we performed a small ablation experiment by training a classical black-box LSTM network, i.e. only training $D$ with all the inputs in $x$, including the power and exogenous temperatures, again with five different random seeds. For comparison purposes, the training and validation losses can be found in Table~\ref{tab: app losses} and the error propagation over the validation set in Figure~\ref{fig: comparison LSTMs}, where one can remark similar to worse performance compared to the proposed PCNNs in Figure~\ref{fig: comparison paper}. This is a strong indication that PCNNs do not lose much expressiveness, even if we constrain the structure to follow some given laws. On the contrary, the linear physics-informed module inside PCNNs seems to give them useful information, as they are able to beat the performance of classical unconstrained LSTMs.
    
    \begin{table}[]
    \centering
    \begin{tabular}{l|c|c|c} \hline
        & Seed & Training loss & Validation loss \\ \hline
        \textit{LSTMs} & $0$ & \remText{$1.31$} \newText{0.57} & \remText{$2.47$} \newText{2.28} \\
        & $1$ & \remText{$1.02$} \newText{0.57} & \remText{$2.56$} \newText{1.92} \\
        & $2$ & \remText{$1.19$} \newText{1.14} & \remText{$2.56$} \newText{2.30} \\ 
        & $3$ & \remText{$1.04$} \newText{$0.97$} & \remText{$2.53$} \newText{$2.22$} \\
        & $4$ & \remText{$1.27$} \newText{$1.00$} & \remText{$2.54$} \newText{$1.77$} \\ \cline{2-4}
        & \textbf{Mean} & $\textbf{\remText{1.16} \newText{0.85}}$ & $\textbf{\remText{2.53} \newText{2.10}}$ \\
        \hline
        \textit{PCNNs} & $0$ & \remText{$1.80$} \newText{1.83} & \remText{$1.96$} \newText{1.93} \\
        & $1$ & \remText{$2.31$} \newText{1.85} & \remText{$1.63$} \newText{1.65} \\
        & $2$ & \remText{$2.35$} \newText{2.06} & \remText{$1.79$} \newText{1.75} \\
        & $3$ & \remText{$2.10$} \newText{2.28} & \remText{$1.83$} \newText{1.73} \\
        & $4$ & \remText{$2.00$} \newText{1.90} & \remText{$1.95$} \newText{1.97} \\
        & $5$ & \remText{$2.00$} \newText{2.38} & \remText{$1.66$} \newText{1.66} \\ \cline{2-4}
        & \textbf{Mean} & $\textbf{\remText{2.09} \newText{2.05}}$ & $\textbf{\remText{1.80} \newText{1.78}}$ \\
        \hline 
    \end{tabular}
    \caption{Comparison training and validation loss for five classical LSTMs and six PCNNs, scaled by $10^{3}$.}
    \label{tab: app losses}
    \end{table}

\section{Deriving PCNNs from classical grey-box models}
    \label{app: contrast}
    
Classical grey-box models generally start from a linear RC model of the following form \cite{sturzenegger2015model}:
\begin{align}
    \dot z(t) &= Az(t) + Bq(t)
\end{align}
where $z$ is the state of the system\footnote{We adopt the unconventional notation $z$ for the state to avoid confusion with the NN inputs $x$ in PCNNs.} and $q$ captures various heat fluxes like heating/cooling inputs, heat losses to the environment and neighboring zones, or heat gains from solar irradiation. After using Euler's discretization, 
we obtain the following discrete-time linear model:
\begin{align}
    z_{k+1} &= Az_k + Bq_k 
\end{align}
Using system identification, one can then identify the parameters $A$ and $B$ 
of this model, yielding a grey-box model.

Traditionally, researchers separate the various heat fluxes in $q$ between \textit{controllable} and \textit{uncontrollable} variables $v$ and $w$, respectively. The former generally captures power inputs in building models, but it could be extended to include blind controls for example. On the other hand, $w$ captures the effect of the sun, the occupants, and other disturbances that cannot be controlled. In the past, both types of heat fluxes have usually been separated linearly, leading to models of the form \cite{sturzenegger2015model}:
\begin{align}
    z_{k+1} &= Az_k + B_vv_k + B_ww_k \label{equ: sturzenegger}
\end{align}
However, while some exogenous factors in $w$ do indeed present a linear impact on the zone temperatures, others are much harder to capture, such as the solar gains or the effect of the occupants. One way to capture more complex phenomena is to introduce bilinear terms coupling $v$ and $w$ in Equation~(\ref{equ: sturzenegger}), as in Sturzenegger et al. \cite{sturzenegger2015model} for example. The natural disadvantage arising from such additional coupling terms in control application is that the subsequent optimization of $v$ gets more complicated.

On the other hand, PCNNs effectively separate the uncontrollable variables into two sets $w^1$ and $w^2$ using prior knowledge on the law of physics in buildings. The former groups variables known to have a linear impact on the zone temperature, the temperature outside and in the neighboring zones. 
$w^2$ then gathers the other inputs with nonlinear effects, 
which computed in a separate disturbance process $\xi$ based on an unknown residual function $m$, which yields a process similar to:
\begin{align}
    \xi_{k+1} &= \xi_k + m(\xi_k, w^2_k) \nonumber \\
    z_{k+1} &= Az_k + B_ug(u_k) + B_{w^1}w^1_k + \xi_{k+1} \label{equ: general RC inspired} 
\end{align}
Note that we introduced a more general form of the controllable inputs $v=g(u)$ since the power inputs $v$ are for example not directly controllable in general. We hence represent them by a function $g(u)$ where $u$ is the true control variable, typically the opening of the valves in the case of radiators. Remarkably, the model presented in Equation~(\ref{equ: general RC inspired}) still retains a linear structure with respect to power inputs $v=g(u)$, which is very well-suited for control applications.

However, PCNNs have yet a slightly different structure: the disturbance model $\xi$ is influencing the state of the system both with and without a lag of one time step, as can be observed in Equation~(\ref{equ: T=D+E}) since $E_{k+1}$ depends on $D_k$. Altogether, we can thus rewrite the equations of the PCNN as follows:
\begin{align}
    \xi_{k+1} &= \xi_k + m(\xi_k, w^2_k) \nonumber \\
    z_{k+1} &= Az_k + B_ug(u_k) + B_{w^1}w^1_k \label{equ: final RC inspired 2} \\
            &\qquad + B_d\xi_k + \xi_{k+1} \nonumber
\end{align}
One can verify that Equations~(\ref{equ: PCNN}) and (\ref{equ: final RC inspired 2}) are equivalent, with:
\begin{align}
    \xi &= D & z &= T \nonumber \\
    w^1 &= \begin{bmatrix} T^{out} & T^{neigh}\end{bmatrix}^T & w^2 &= x \nonumber \\
    A &= 1-b-c & B_u &= a \nonumber \\
    B_{w^1} &= \begin{bmatrix} -b & -c \end{bmatrix} & B_d &= -b-c \nonumber
\end{align}

\end{document}